\documentclass[preprint,3p,12pt]{elsarticle}
\biboptions{sort&compress}

\usepackage{amsmath}
\usepackage{amsfonts}
\usepackage{amssymb}
\usepackage{amsthm}
\usepackage{enumitem}
\usepackage{algorithm}
\usepackage{algpseudocode}
\usepackage{xcolor}

\usepackage{geometry}
\usepackage{float}
\usepackage{graphicx}
\usepackage{subcaption}
\usepackage{booktabs}

\usepackage{aliascnt}
\usepackage[hyphens]{url}
\usepackage[colorlinks=true, linkcolor=blue, citecolor=red, urlcolor=blue]{hyperref}
\usepackage[capitalize, nameinlink]{cleveref} % capitalize 参数强制首字母大写

\theoremstyle{remark}

\newtheorem{theorem}{Theorem}[section]
\newaliascnt{lemma}{theorem}
\newtheorem{lemma}[lemma]{Lemma}
\aliascntresetthe{lemma}
\crefname{theorem}{Theorem}{Theorems}
\Crefname{theorem}{Theorem}{Theorems}
\crefname{lemma}{Lemma}{Lemmas}
\Crefname{lemma}{Lemma}{Lemmas}
\theoremstyle{definition}

\begin{document}

\begin{frontmatter}

\title{FDM+: A Framework for Constrained Control of the 2D Radiative Transport Equation}

\author[label1,label3]{Hualing Zhong}
\author[label1]{Shuai Wang}
\author[label1]{Hui Xie\corref{cor1}}
\author[label1,label2]{Heng Yong}
\cortext[cor1]{Corresponding author. Email address: \href{mailto:xiehui@lsec.cc.ac.cn}{xiehui@lsec.cc.ac.cn}}

\affiliation[label1]{organization={Institute of Applied Physics and Computational Mathematics},
            city={Beijing},
            postcode={100094},
            country={China}}

\affiliation[label2]{organization={National Key Laboratory of Computational Physics},
            city={Beijing},
            postcode={100083},
            country={China}}

\affiliation[label3]{organization={Graduate School of China Academy of Engineering Physics},
            city={Beijing},
            postcode={100094},
            country={China}}

\begin{abstract}
Applying the frozen diffusion model (FDM) \cite{zhong2025constrained} to the constrained control of the 2D radiative transport equation presents several challenges. These include the dimensional difference between control variables and state functions, the resolution mismatch between traditional numerical solvers and deep learning models, and solution non-uniqueness. To address these obstacles, we propose a novel approach, FDM+, which includes three key improvements. Firstly, a decoupled prediction architecture is introduced to address the dimensional heterogeneity. Secondly, an efficient parameterization strategy resolves the resolution mismatch. Meanwhile, we derive a total error bound to analyze how the parameterization error and the model estimation error jointly affect the performance. Thirdly, the conditional distribution learned by the diffusion model is leveraged to perform design space exploration for multi-solution problems. A series of numerical experiments demonstrate that FDM+ outperforms existing methods in reconstruction accuracy and design flexibility.
\end{abstract}

\begin{keyword}
Frozen diffusion model, Radiative transport equation, Constrained control, Decoupled prediction, Control parameterization, Design space exploration
\end{keyword}

\end{frontmatter}

%% Main text
\section{Introduction}
\label{sec1}
Radiative transport is a fundamental physical process in fields such as astrophysics \cite{castor2007astrophysical}, inertial confinement fusion \cite{atzeni2005fluid}, and high-temperature fluid dynamics \cite{viskanta1998overview}. The process is typically governed by the radiative transport equations (RTEs), for which analytical solutions are often unattainable due to high dimensionality, strong nonlinearity, and multiscale features. These complexities make designing solutions to achieve specific objectives an important challenge, especially in applications demanding high safety and precision, such as nuclear reactor shielding \cite{santoro2000radiation} and medical optical tomography \cite{arridge2009optical}. This is a class of constrained control problems: tuning variables, such as medium properties or boundary conditions, to guide a complex system's solution towards a desired state \cite{egger2017numerical, li2023monte, ulbrich2018introduction, kovtanyuk2014theoretical}. Traditional optimal control methods, such as adjoint-based gradient descent, often struggle with the highly non-convex loss landscapes of these problems. Similarly, while reinforcement learning approaches have been explored, they face data inefficiency and reward sparsity when navigating the high-dimensional RTEs. Solving such optimization problems numerically typically relies on an iterative framework that requires repeatedly solving the forward RTEs, which is costly. Consequently, conventional methods like the finite difference \cite{castor2007radiation}, finite element \cite{kanschat2009numerical, reed1973triangular}, and finite volume \cite{chai1994finite} methods, while effective for forward simulations, prove impractical for control tasks and rapid exploration of the design space \cite{arridge2009optical, klose1999iterative, saratoon2013gradient}.

To overcome the computational bottlenecks, researchers have increasingly turned to deep learning. This paradigm offers a fundamental shift in approach: instead of repeatedly solving the RTEs to find a solution, deep learning models are trained to directly approximate the complex mapping from a desired physical state to the corresponding control variables. A variety of architectures have been investigated for this purpose, ranging from data-driven models like feedforward and convolutional neural networks (FNNs, CNNs) \cite{liu2019neural, han2018solving, zha2022convolution, long2019pde} to physics-informed neural networks (PINNs) \cite{raissi2019physics, xie2024conservative, liu2024discontinuity, mei2024solving} that embed the governing equations into the training process. Besides, operator learning frameworks, such as DeepONet \cite{lu2021learning} and Fourier neural operator (FNO) \cite{li2021fourier}, have also shown significant promise in learning mappings between infinite-dimensional function spaces. Despite these advances \cite{wang2021learning, shi2024physics, karniadakis2021physics}, deterministic deep learning methods suffer from three major limitations: requiring massive training data, insufficient numerical accuracy for the RTEs, and the inability to handle multi-solution problems \cite{karniadakis2021physics, zhong2025constrained}.

To address these limitations, researchers have turned to diffusion models, which have achieved state-of-the-art results in generative modeling \cite{ho2020denoising, song2021scorebased, song2021denoising, lipman2023flow, cao2024survey} and are increasingly popular in the scientific computing \cite{lienen2024zero, holzschuh2023solving, li2024learning, shu2023physics, ruhling2024dyffusion, shysheya2024conditional}. For constrained control problems, The greatest advantage of diffusion model is the ability to condition the generation process on known information, providing a mechanism for enforcing giving constraints. However, current conditioning techniques still present some drawbacks, which can be categorized based on where the conditioning is applied. Firstly, \textit{conditioning during the denoising process} often relies on classifier or classifier-free guidance. These methods introduce inefficient auxiliary models or sensitive, hard-to-tune hyperparameters \cite{dhariwal2021classifier, ho2021classifier-free, huang2024diffusionpde, wei2024diffphycon}. The second approach is to \textit{condition during the training process} by embedding constraints into the loss function, but this creates an intractable challenge of balancing objective terms \cite{bastek2024physics}. The third strategy involves \textit{conditioning the noising process}, for example, by applying different noise schedules to known and unknown data. This, however, may corrupt the integrity of the known conditions \cite{liu2024residual, kohl2024benchmarking}. These challenges underscore the need for a more direct and efficient approach to conditional generation for constrained control problems.

Our prior work introduced the frozen diffusion model (FDM) to address these conditioning drawbacks for the constrained control of one-dimensional (1D) RTE \cite{zhong2025constrained}. The core idea of FDM is to freeze the known conditions, providing a direct conditional generation mechanism. This framework proved high precision and data-efficient for 1D problems. However, applying FDM to the constrained control problem for the 2D RTE introduces several challenges. A primary challenge is the \textbf{dimensional heterogeneity}. This issue arises because the model must learn the joint distribution between a low-dimensional control function and a high-dimensional state field, which poses a major architectural challenge for neural network modeling. Another fundamental issue is the \textbf{scale mismatch}, because the temporal resolutions required by high-fidelity numerical solvers far exceed the practical sequence length limitations of diffusion model, making the direct generation of full-resolution solutions computationally infeasible. Furthermore, the problem of \textbf{non-uniqueness}, while present in 1D, becomes more severe in high dimensions. For a 2D problem, a vast solution space of different control functions can satisfy the same objective, raising the critical question of how to effectively explore this space for an optimal design.

This paper presents a more complete framework, FDM+, which introduces three key improvements for the constrained control of 2D RTE. \textbf{Firstly, to resolve the dimensional heterogeneity, we propose a novel decoupled prediction architecture.} This architecture uses parallel pathways within the network to independently process and predict the low-dimensional control variables and the high-dimensional state field, allowing the model to learn their distinct structures effectively. \textbf{Secondly, to overcome the scale mismatch, we develop an efficient control parameterization strategy.} Instead of generating a long, high-resolution control sequence, our model learns a set of low-dimensional control nodes. Furthermore, we present a total error bound that incorporates both the parameterization and estimation errors. \textbf{Finally, we leverage the stochasticity of diffusion model to explore the design space.} By sampling multiple times from different initial noise vectors, FDM+ generates an ensemble of high-quality, diverse solutions that satisfy the constraints. These valid candidates enables post-hoc selection of the optimal design.

The remainder of this paper is organized as follows. \cref{sec2} establishes the mathematical preliminaries, formulating the 2D RTE control problem, reviewing the FDM, and detailing the challenges of applying FDM to 2D RTE. \cref{sec3} introduces our FDM+ method, including the decoupled prediction architecture, the control parameterization, and the design space exploration. \cref{sec4} validates the framework via numerical experiments, including ablation study on a single medium, baseline comparisons across multiple media, and multi-solution exploration under the diffusion limit of radiative transport. Finally, \cref{sec5} concludes the paper.

\section{Preliminaries}
\label{sec2}

\subsection{2D Radiative Transport Equation and Constrained Control}
\label{subsec2.1}
The radiative transport equation (RTE) system describes the evolution of radiation intensity in space, angle, time, and frequency. The governing equation is:
\begin{equation}
\left\{
    \begin{aligned}
        & \frac{1}{c}\frac{\partial I}{\partial t}+\boldsymbol{\Omega}\cdot\nabla I = \sigma_a(B(\nu,T)-I) + Q, \\
        & C_{v}\frac{\partial T}{\partial t}=\int_{S^{2}}\int_{0}^{\infty}\sigma_a(I-B(\nu,T))dvd\boldsymbol{\Omega}.
    \end{aligned}
\right.
\end{equation}
Here, $I(\boldsymbol{x},t,\nu,\boldsymbol{\Omega})$ is the angular radiation intensity, dependent on the spatial coordinate $\boldsymbol{x}$, time $t$, frequency $\nu$, and direction $\boldsymbol{\Omega}$. The constant $c$ is the speed of light, $T$ is the temperature of the background medium, $\sigma_a$ is the absorption coefficient, $C_v$ is the volume heat capacity, which is an integral operator dependent on the radiation intensity $I$, and $B(\nu,T)$ is the Planck function:
\begin{equation}
    B(\nu,T)=\frac{2h\nu^{3}}{c^{2}}\frac{1}{e^{h\nu/k_bT}-1},
\end{equation}
where $h$ is the Planck constant and $k_b$ is the Boltzmann constant.

In this work, we consider the grey RTE in a 2D Cartesian domain $D \subset \mathbb{R}^2$, which is governed by:
\begin{equation}
\label{eq3}
\left\{
    \begin{aligned}
        & \frac{1}{c}\frac{\partial I}{\partial t} + \boldsymbol{\Omega}\cdot\nabla I = \sigma_a\left(\frac{1}{4\pi}acT^4 - I\right), & (x,y) \in D, \ t > 0, \\
        & C_v\frac{\partial T}{\partial t} = \sigma_a\left(\int_0^{2\pi}\int_0^\pi I\sin\theta d\theta d\varphi - acT^4\right), & (x,y) \in D, \ t > 0, \\
        & I(x, y, \theta, \varphi, t > 0) = S(t), & (x,y) \in \partial D, \ \boldsymbol{\Omega}\cdot\vec{n} < 0, \\
        & I(x, y, \theta, \varphi, t=0) = \frac{1}{4\pi}acT(x, y, t=0)^4, & (x,y) \in D, \\
        & T(x, y, t=0) = T_0, & (x,y) \in D.
    \end{aligned}
\right.
\end{equation}
In this system, $\boldsymbol{\Omega}$ is parameterized by angles $(\theta, \varphi)$, leading to $\boldsymbol{\Omega}\cdot\nabla I = \sin\theta\cos\varphi\frac{\partial I}{\partial x} + \sin\theta\sin\varphi\frac{\partial I}{\partial y}$. We consider a temperature-dependent opacity of the form $\sigma_a(T) = \sigma / T^3$, where $\sigma$ is an intrinsic opacity constant.

Based on the physical model above, this paper aims to solve the constrained control problem: for a given target radiation state in a specific spatial region $D^* \subset D$, find the unknown inflow boundary condition $S(t)$ such that the solution driven by this boundary condition approximates the target state in $D^*$ as closely as possible. This problem can be formally defined as minimizing the following objective:
\begin{equation}\label{eq:objective}
    \text{find } S^*(t) = \underset{S(t)}{\mathrm{argmin}} \left\| \rho(x,y,t) - \rho^*(x,y,t) \right\|, \quad (x,y) \in D^* \subset D,
\end{equation}
where $\rho$ is the scalar flux, defined as the integral of the radiation intensity over all solid angles:
\begin{equation}
    \rho = \frac{1}{4\pi}\int I d\boldsymbol{\Omega}.
\end{equation}
Note that the scalar flux $\rho$ is directly proportional to the radiation energy density $E_r$ via $E_r = \frac{4\pi}{c}\rho$, making the control of $\rho$ physically equivalent to controlling the radiation energy.

\subsection{Review of the Frozen Diffusion Model}
\label{subsec2.2}
Diffusion models learn a data distribution through a paired forward noising process and a reverse denoising process \cite{ho2020denoising, luo2022understanding}. In the standard forward process, Gaussian noise is progressively added to the initial data $\mathbf{X}_0$ over $K$ discrete steps. The noisy data $\mathbf{X}_k$ at any step $k$ can be sampled directly from $\mathbf{X}_0$:
\begin{equation}
    \mathbf{X}_k = \sqrt{\bar{\alpha}_k}\mathbf{X}_0 + \sqrt{1-\bar{\alpha}_k}\boldsymbol{\epsilon}, \quad  \boldsymbol{\epsilon} \sim \mathcal{N}(0, \mathbf{I}),
\end{equation}
where $\{\bar{\alpha}_k\}_{k=1}^K$ are predefined noise schedule coefficients.

The model trains a neural network $\boldsymbol{\epsilon}_\theta(\mathbf{X}_k, k)$ to predict the noise $\boldsymbol{\epsilon}$ added during the forward process by minimizing the mean squared error:
\begin{equation}
    \mathcal{L} = \mathbb{E}_{k, \mathbf{X}_0, \boldsymbol{\epsilon}} \left[ ||\boldsymbol{\epsilon} - \boldsymbol{\epsilon}_\theta(\mathbf{X}_k, k)||^2 \right].
\end{equation}

During generation, the model starts from pure noise $\mathbf{X}_K \sim \mathcal{N}(0, \mathbf{I})$ and iteratively denoises it. Each step, from $k$ to $k-1$, computes the denoised sample $\mathbf{X}_{k-1}$ as follows:
\begin{equation}
    \mathbf{X}_{k-1} = \frac{1}{\sqrt{\alpha_k}}\left(\mathbf{X}_k - \frac{1-\alpha_k}{\sqrt{1-\bar{\alpha}_k}}\boldsymbol{\epsilon}_\theta(\mathbf{X}_k, k)\right) + \sqrt{\frac{1-\bar{\alpha}_{k-1}}{1-\bar{\alpha}_k} (1-\alpha_k)} \cdot \mathbf{z}
\end{equation}
where $\mathbf{z} \sim \mathcal{N}(0, \mathbf{I})$ for $k>1$ (and $\mathbf{z}=0$ for $k=1$).

When addressing the constrained control problems, the data $\mathbf{X}_k$ typically consists of two parts: the unknown control variables to be solved for and the known target state. Based on this insight, the FDM \cite{zhong2025constrained} framework freezes the known conditions. Specifically, we only add noise to the unknown parts of the data while keeping the known conditions unchanged. This is achieved mathematically using a mask tensor $\mathbf{M}$ that identifies the locations of known information (1 for known, 0 for unknown). The FDM noising process is thus modified as:
\begin{equation}
    \mathbf{X}_k = (\mathbf{I}-\mathbf{M}) \odot (\sqrt{\bar{\alpha}_k}\mathbf{X}_0 + \sqrt{1-\bar{\alpha}_k}\boldsymbol{\epsilon}) + \mathbf{M} \odot \mathbf{X}_0,
\end{equation}
where $\odot$ denotes the element-wise product and $\mathbf{I}$ is a tensor of ones.

\begin{figure}[H]
    \centering
    \includegraphics[width=0.8\textwidth]{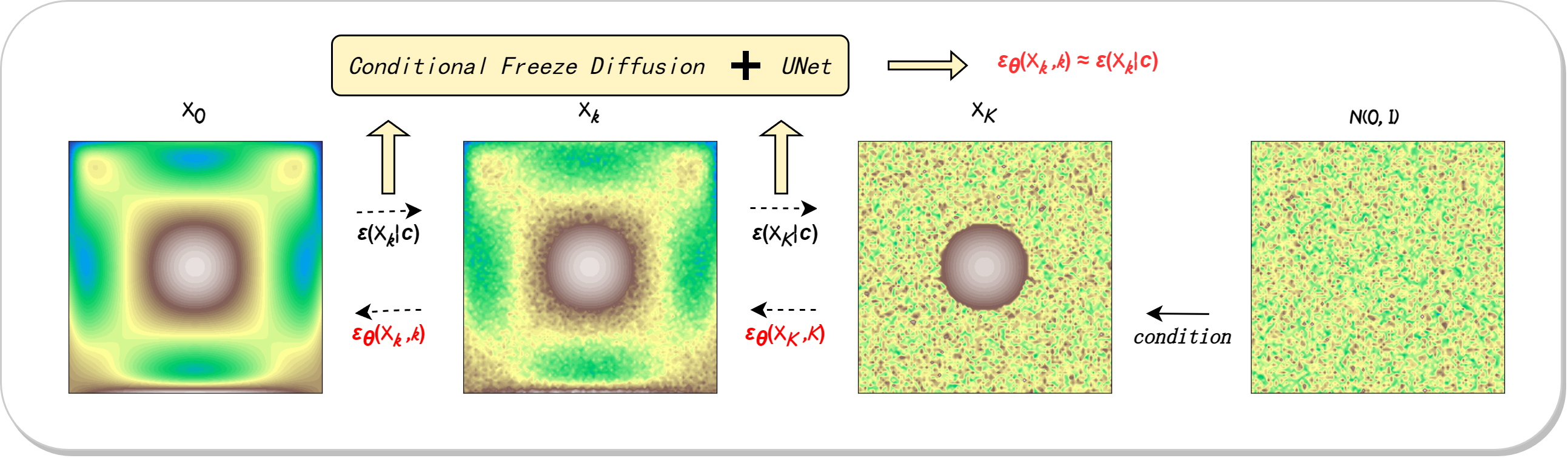}
    \caption{The framework of the frozen diffusion model (FDM). The central part is frozen as the known conditions, and $\varepsilon(\mathbf{X}_{k}|c)$ represents the conditional noise.}
    \label{fig:framework}
\end{figure}

As illustrated in \cref{fig:framework}, the network receives the partially noisy and clean state $\mathbf{X}_k$ during training to predict the noise $\boldsymbol{\epsilon}$ added to the unknown regions. In each step of the reverse denoising process, the known conditions serve as a powerful guide, ensuring that the generated data matches the target state. In our prior 1D work \cite{zhong2025constrained}, we proved that this freezing modification preserves the Markov property of both the forward and reverse processes, which ensures the theoretical rigor.

\subsection{Challenges in Applying FDM to 2D Problem}
\label{subsec2.3}
While FDM successfully solves the constrained control of 1D RTE, scaling it to 2D faces some difficulties. For the 2D RTE in this article, the boundary control $S(t)$ is a low-dimensional temporal signal, whereas the target state $\rho(x, y, t)$ is a high-dimensional spatio-temporal field. Jointly modeling both variables requires expanding the low-dimensional control across spatial dimensions, which causes the diffusion model to predict low-quality control signals contaminated by spatial noise. Meanwhile, for the initial value at low temperature ($T_0 \ll 1$), high-fidelity solvers require ultra-fine temporal grids to resolve stiff physics, forcing the control variable $S(t)$ to be discretized into thousands of steps. Directly generating such a long, high-resolution sequence for $S(t)$ via diffusion models leads to prohibitive computational costs and error accumulation. Finally, the complexity of 2D transport allows multiple completely distinct boundary conditions $S(t)$ to generate the identical localized target state $\rho(x, y, t)$. This forces FDM to address the optimal design of control. In short, these challenges of dimensions, scales, and non-uniqueness render the FDM ineffective, motivating the new framework of FDM+.

\section{Methodology}
\label{sec3}

\subsection{Decoupled Prediction Architecture for Dimensional Heterogeneity}
\label{subsec3.1}

\begin{figure}[H]
    \centering
    \includegraphics[width=0.8\textwidth]{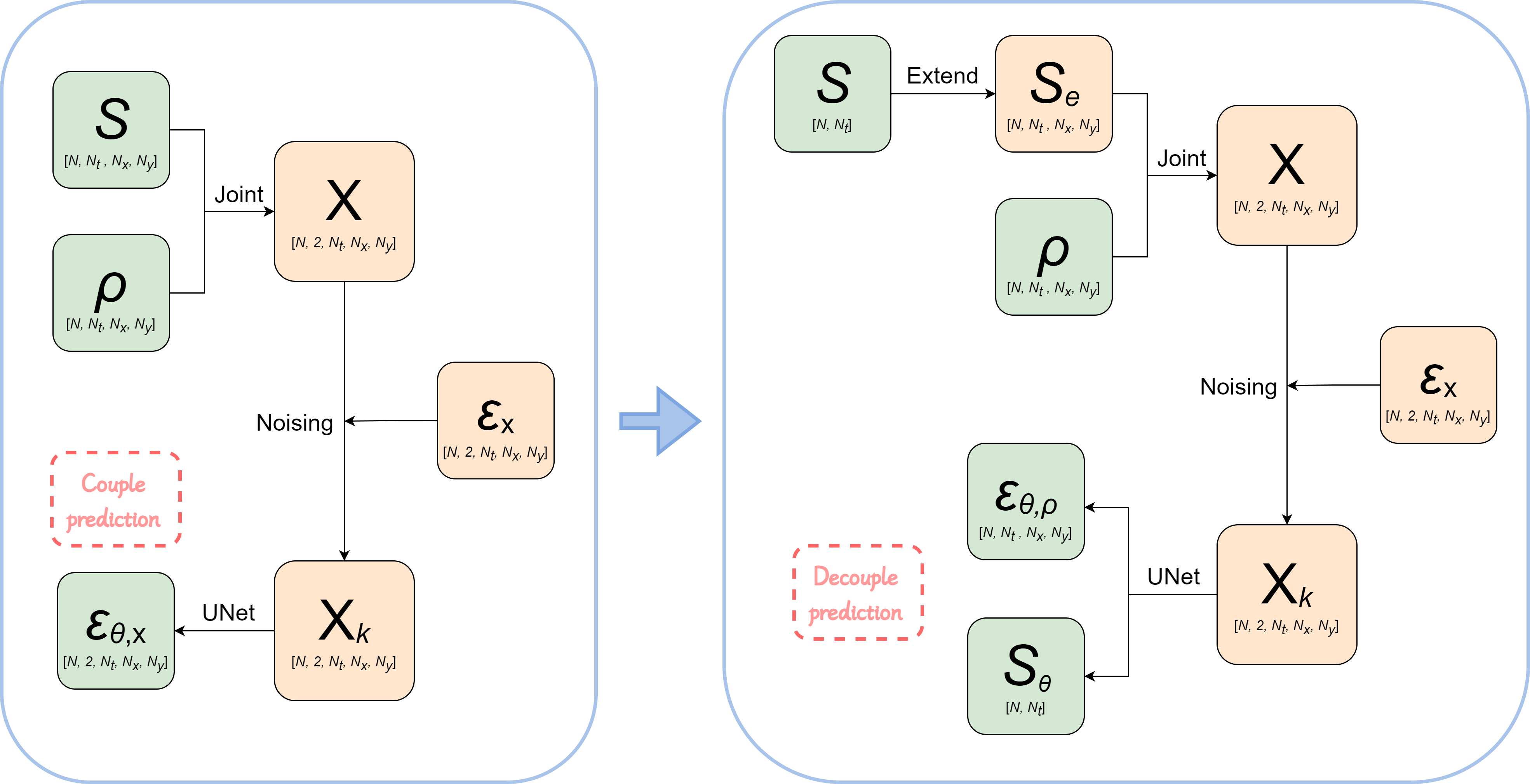}
    \caption{A comparison of the standard prediction objective and our proposed decoupled prediction Architecture. \textbf{(Left)} The standard approach trains a UNet \cite{ronneberger2015u} to predict noise $\boldsymbol{\epsilon}_{\theta,\mathbf{X}}$ for the entire joint state. \textbf{(Right)} Our decoupled mechanism employs a hybrid objective: the network predicts the noise $\boldsymbol{\epsilon}_{\theta,\rho}$ for the high-dimensional field $\rho$, but directly predicts the original, clean values $S_{\theta}$ for the low-dimensional control $S$. Bracketed notation (e.g., $[N, N_t, N_x, N_y]$) denotes the tensor shape, where $N$ is the batch size, $N_t$ represents the temporal dimension, and $N_x, N_y$ represent the spatial dimensions.}
    \label{fig:decouple}
\end{figure}

To resolve dimensional heterogeneity, we modify the prediction objective of FDM, proposing a new architecture of "Joint Modeling, Decoupled Prediction", as depicted in \cref{fig:decouple}. For the high-dimensional flux field with complex spatial textures, we retain the standard noise prediction objective $\boldsymbol{\epsilon}_{\rho}$. In contrast, for the spatially expanded control variables, we have the network directly predict its original, noise-free values $S$. This hybrid prediction objective is realized through a weighted loss function:
\begin{equation}
    L = \mathcal{L}(\boldsymbol{\epsilon}_{\rho}, \boldsymbol{\epsilon}_{\theta, \rho}) + \lambda \cdot \mathcal{L}(S, S_{\theta}),
\end{equation}
where $\lambda$ is a weighting hyperparameter. The first term is the noise prediction loss for the flux and the second is the direct value prediction loss for the boundary control. We establish a theoretical justification for this hybrid objective below.

\begin{theorem}[Equivalence of hybrid prediction]
\label{theorem1}
Let the noisy state be $\mathbf{X}_k = [\rho_k, S_k]$. Let the model $N_{\theta}(\mathbf{X}_k, k)$ produce a hybrid output $N_{\theta} = [\boldsymbol{\epsilon}_{\theta,\rho}, S_{\theta}]$, where $\boldsymbol{\epsilon}_{\theta,\rho}$ is the predicted noise for component $\rho$, and $S_{\theta}$ is the direct prediction of the clean state for component $S$. This parameterization is equivalent to the standard noise prediction model.
\end{theorem}
\begin{proof}
The standard diffusion model recovery relationship is:
\begin{equation}\label{eq:recon_eng}
    \mathbf{X}_0 = \frac{1}{\sqrt{\bar{\alpha}_k}}(\mathbf{X}_k - \sqrt{1-\bar{\alpha}_k}\boldsymbol{\epsilon}) \quad \Leftrightarrow \quad \boldsymbol{\epsilon} = \frac{1}{\sqrt{1-\bar{\alpha}_k}}(\mathbf{X}_k - \sqrt{\bar{\alpha}_k}\mathbf{X}_0).
\end{equation}
To prove equivalence, we must show that the hybrid output $[\boldsymbol{\epsilon}_{\theta,\rho}, S_{\theta}]$ uniquely determines an equivalent full noise prediction $\hat{\boldsymbol{\epsilon}} = [\hat{\boldsymbol{\epsilon}}_\rho, \hat{\boldsymbol{\epsilon}}_S]$.

First, we define the estimated clean state $\hat{\mathbf{X}}_0 = [\hat{\rho}_0, \hat{S}_0]$ from the model's hybrid output. The component $\hat{S}_0$ is given directly, while $\hat{\rho}_0$ is calculated using Eq.~\eqref{eq:recon_eng}:
\begin{align}
    \hat{S}_0 &:= S_{\theta}, \\
    \hat{\rho}_0 &:= \frac{1}{\sqrt{\bar{\alpha}_k}}(\rho_k - \sqrt{1-\bar{\alpha}_k}\boldsymbol{\epsilon}_{\theta,\rho}).
\end{align}
Since $S_{\theta}$ and $\boldsymbol{\epsilon}_{\theta,\rho}$ are direct outputs of the neural network, the estimate $\hat{\mathbf{X}}_0$ is uniquely determined.

Next, we construct the equivalent full noise prediction $\hat{\boldsymbol{\epsilon}}$ using this unique $\hat{\mathbf{X}}_0$ and Eq.~\eqref{eq:recon_eng}. The component $\hat{\boldsymbol{\epsilon}}_\rho$ is simply the model's output, while $\hat{\boldsymbol{\epsilon}}_S$ is derived from $\hat{S}_0$:
\begin{align}
    \hat{\boldsymbol{\epsilon}}_\rho &:= \boldsymbol{\epsilon}_{\theta,\rho}, \\
    \hat{\boldsymbol{\epsilon}}_S &:= \frac{1}{\sqrt{1-\bar{\alpha}_k}}(S_k - \sqrt{\bar{\alpha}_k}\hat{S}_0) = \frac{1}{\sqrt{1-\bar{\alpha}_k}}(S_k - \sqrt{\bar{\alpha}_k}S_{\theta}).
\end{align}
This demonstrates a unique, one-to-one mapping from the hybrid output $[\boldsymbol{\epsilon}_{\theta,\rho}, S_{\theta}]$ to a full noise prediction $\hat{\boldsymbol{\epsilon}}$, ensuring a well-defined reverse process.
\end{proof}

Although \cref{theorem1} guarantees mathematical equivalence, the decoupled formulation offers critical advantages in predictive performance. In standard coupled framework, predicting the spatially expanded control signal requires collapsing the tensor back to a low-dimensional form via aggregation operations. This process inevitably introduces spatial noise. By contrast, directly predicting the clean control through independent output modules completely avoids such noise interference. As shown in the ablation study (\cref{subsec4.1}), this mechanism significantly enhances reconstruction accuracy.

\subsection{Efficient Control Parameterization for Resolution Mismatch}
\label{subsec3.2}

Directly generating high-resolution temporal control sequences using diffusion models is computationally expensive and suffers from the curse of dimensionality. We therefore parameterize the control function using a low-dimensional node vector $\mathbf{s}\in\mathbb{R}^n$. Since the numerical solver requires a high-resolution boundary input, the continuous control $S_{\mathrm{interp}}(t;\mathbf{s})$ is interpolated by the predicted nodes. The diffusion model is then trained to learn the conditional node distribution $p_n(\mathbf{s}\mid\rho^*)$, where $\rho^*$ is the conditions and the subscript $n$ indicates the node dimension.

Increasing $n$ improves the temporal representation of the control, but also increases the dimension of the distribution estimated from a finite training set. We quantify these two effects below. Throughout this subsection, $\rho^*$ is fixed.

Let $T_f>0$ denote the terminal time of the control process. For each integer $n\geq2$, let
\begin{equation}
t_i=\frac{(i-1)T_f}{n-1},
\qquad i=1,\ldots,n,
\label{eq:uniform_node_grid}
\end{equation}
and define the node sampling map and piecewise-linear interpolation
operator by
\begin{equation}
E_nS
=
\bigl(S(t_1),\ldots,S(t_n)\bigr),
\qquad
S_{\mathrm{interp}}(\cdot;\mathbf{s})=I_n\mathbf{s}.
\label{eq:node_vector_interpolant}
\end{equation}
Let $\mathcal P(\cdot\mid\rho^*)$ denote the conditional distribution of $S$, and $p_n(\cdot\mid\rho^*)$ be the distribution of $\mathbf{s}=E_nS$. $\mathcal D_{n,M}$ denotes a training set of $M$ samples drawn from $p_n(\cdot\mid\rho^*)$.

\begin{lemma}[Interpolation Error for Lipschitz Continuous]
\label{lem:lipschitz_interpolation}
Let $S:[0,T_f]\to[0,S_{\max}]$ be $L$-Lipschitz continuous. Then its
piecewise-linear interpolant on the grid
Eq.~\eqref{eq:uniform_node_grid} satisfies
\begin{equation}
\left\|
S-I_nE_nS
\right\|_{L_2([0,T_f])}
\leq
\frac{LT^{3/2}_f}{2(n-1)}.
\label{eq:lipschitz_interpolation_error}
\end{equation}
\end{lemma}

The proof of \cref{lem:lipschitz_interpolation} is deferred to \ref{app:lipschitz_interpolation}.

\begin{lemma}[TV Estimation Rate for Diffusion Models
\cite{oko2023diffusion}]
\label{lem:diffusion_tv}
Suppose that, after an invertible affine rescaling to
$[-1,1]^n$, the conditional node density $p_n(\cdot\mid\rho^*)$ satisfies Assumptions~2.4--2.6 of \cite{oko2023diffusion}, and that the diffusion estimator is constructed under the conditions of Theorem~5.1 therein. Then, for
sufficiently large $M$,
\begin{equation}
\mathbb E_{\mathcal D_{n,M}}
D_{\mathrm{TV}}\!\left(
\hat p_{n,M}(\cdot\mid\rho^*),
p_n(\cdot\mid\rho^*)
\right)
\leq
C_{\mathrm{TV},n}
M^{-\frac{\beta}{2\beta+n}}
(\log M)^8,
\label{eq:diffusion_tv_rate}
\end{equation}
where $\beta>0$ is the Besov smoothness and $C_{\mathrm{TV},n}>0$ is independent of $M$.
\end{lemma}

\begin{lemma}[Maximal Coupling \cite{Lindvall2002couple}]
\label{lem:maximal_coupling}
Let $\mu$ and $\nu$ be Borel probability measures on
$\mathbb{R}^n$. There exists a coupling $(X,Y)$ of $\mu$ and $\nu$
such that
\begin{equation}
\mathbb P(X\neq Y)
=
D_{\mathrm{TV}}(\mu,\nu),
\qquad
D_{\mathrm{TV}}(\mu,\nu)
=
\sup_A|\mu(A)-\nu(A)|.
\label{eq:maximal_coupling_identity}
\end{equation}
\end{lemma}

Let $\hat p_{n,M}(\cdot\mid\rho^*)$ denote the node distribution estimated from a training set $\mathcal D_{n,M}$ of size $M$. Generated nodes are constrained componentwise to $[0,S_{\max}]$. If necessary, this can be enforced by clipping. For each realization of $\mathcal D_{n,M}$, couple $S\sim\mathcal P(\cdot\mid\rho^*)$ and
$\hat{\mathbf{s}}\sim\hat p_{n,M}(\cdot\mid\rho^*)$ such that $(E_nS,\hat{\mathbf{s}})$ is a maximal coupling of
$p_n(\cdot\mid\rho^*)$ and $\hat p_{n,M}(\cdot\mid\rho^*)$. Define the total error by
\begin{equation}
\mathcal R_{\mathrm{total}}(n\mid\rho^*)
=
\mathbb E_{\mathcal D_{n,M}}
\mathbb E
\left[
\left\|
S-I_n\hat{\mathbf{s}}
\right\|_{L_2([0,T_f])}
\right].
\label{eq:total_reconstruction_error}
\end{equation}

\begin{theorem}[Total Error Bound]
\label{thm:reconstruction_error_bound}
Suppose that
$S\sim\mathcal P(\cdot\mid\rho^*)$ takes values in
$[0,S_{\max}]$ and is $L$-Lipschitz continuous almost surely.
Under the assumptions of \cref{lem:diffusion_tv},
\begin{equation}
\mathcal R_{\mathrm{total}}(n\mid\rho^*)
\leq
\frac{LT^{3/2}_f}{2(n-1)}
+
C_{\mathrm{TV},n}S_{\max}\sqrt{T_f}\,
M^{-\frac{\beta}{2\beta+n}}
(\log M)^8.
\label{eq:total_error_bound}
\end{equation}
\end{theorem}

\begin{proof}
Let $\mathbf{s}=E_nS$. By the triangle inequality,
\begin{equation}
\left\|
S-I_n\hat{\mathbf{s}}
\right\|_{L_2}
\leq
\left\|
S-I_n\mathbf{s}
\right\|_{L_2}
+
\left\|
I_n\mathbf{s}-I_n\hat{\mathbf{s}}
\right\|_{L_2}.
\label{eq:reconstruction_error_decomposition}
\end{equation}
The first term is bounded by
\cref{lem:lipschitz_interpolation}. For the second term, both
interpolants take values in $[0,S_{\max}]$, and hence
\begin{equation}
\left\|
I_n\mathbf{s}-I_n\hat{\mathbf{s}}
\right\|_{L_2([0,T_f])}
\leq
S_{\max}\sqrt{T_f}\,
\mathbb I_{\{\mathbf{s}\neq\hat{\mathbf{s}}\}}.
\label{eq:coupled_interpolation_bound}
\end{equation}
By the maximal coupling property,
\begin{equation}
\mathbb E
\left[
\left\|
I_n\mathbf{s}-I_n\hat{\mathbf{s}}
\right\|_{L_2([0,T_f])}
\right]
\leq
S_{\max}\sqrt{T_f}\,
D_{\mathrm{TV}}\!\left(
p_n(\cdot\mid\rho^*),
\hat p_{n,M}(\cdot\mid\rho^*)
\right).
\label{eq:generative_error_tv_bound}
\end{equation}
Taking expectation over $\mathcal D_{n,M}$ and applying
\cref{lem:diffusion_tv} proves Eq.~\eqref{eq:total_error_bound}.
\end{proof}

Although the dependence of $C_{\mathrm{TV},n}$ on $n$ is not
explicitly characterized, the dimension-dependent factor
$M^{-\frac{\beta}{2\beta+n}}$ is strictly increasing in $n$. This observation does not establish monotonicity of the total error or the existence of an optimal node count. Nevertheless, it shows that increasing $n$ is not guaranteed to improve the total error and motivates investigating whether an intermediate node count performs best. This behavior is examined in the numerical experiments (\cref{subsec4.1}).

\subsection{Design Space Exploration}
\label{subsec3.3}
The constrained control problems for 2D RTE are highly ill-posed, resulting in severe non-uniqueness. Traditional approaches often enforce a single solution through regularization \cite{engl1996regularization}, which unduly constrains engineering design options. In contrast, FDM+ operates as a generative model that learns the complete conditional distribution $p(S, \rho \mid \rho^*)$ over the feasible manifold. By sampling with different initial noise vectors, FDM+ generates a diverse ensemble of distinct control curves $S_i(t)$. When validated using the traditional high-fidelity numerical solver, all these candidates successfully satisfy the primary target $\rho^*$. This multi-solution capability enables a posteriori evaluation based on secondary objectives, such as minimizing boundary energy, peak power, or spatio-temporal gradient variations. The selection workflow is formalized in \cref{alg:design_explore}.

\begin{algorithm}[H]
\caption{Posteriori multi-objective design selection via batch sampling}
\label{alg:design_explore}
\begin{algorithmic}[1]
\Require Trained FDM+, control target $\rho^*$, ensemble size $P$, secondary objectives $\mathcal{J}$, primary objective $J_{\text{primary}}$, error threshold $\epsilon_{\text{tol}}$.
\Ensure A filtered design portfolio $\mathcal{H}$ of high-fidelity solutions.
\State Sample ensemble: $\{\hat{S}_1, \dots, \hat{S}_P\} \leftarrow \text{FDM+}(\cdot | \rho^*, P)$.
\State Initialize portfolio: $\mathcal{H} \leftarrow \emptyset$.
\For{$i = 1$ to $P$}
    \State Compute primary error: $e_i \leftarrow J_{\text{primary}}(\hat{S}_i, \rho^*)$. \Comment{Requires forward solve}
    \If{$e_i \le \epsilon_{\text{tol}}$}
        \State Initialize performance vector: $\mathbf{v}_i \leftarrow \emptyset$.
        \For{$k = 1$ to $m$}
            \State $v_{i,k} \leftarrow J_k(\hat{S}_i)$.
            \State $\mathbf{v}_i \leftarrow \mathbf{v}_i \cup \{v_{i,k}\}$.
        \EndFor
        \State Add to portfolio: $\mathcal{H} \leftarrow \mathcal{H} \cup \{(\hat{S}_i, e_i, \mathbf{v}_i)\}$.
    \EndIf
\EndFor
\State \Return The filtered design portfolio $\mathcal{H}$.
\end{algorithmic}
\end{algorithm}

\section{Numerical Experiments}
\label{sec4}

We evaluate the FDM+ framework through a series of constrained control problems. First, a node-count scan and an ablation study in a homogeneous medium assess the proposed parameterization and the decoupled prediction. Second, experiments in two heterogeneous media compare FDM+ with other methods. Third, we explore design spaces under the 2D RTE diffusion limit. The data generation procedures are reported in \ref{app:data_generation}, and the network and training settings are given in \ref{app:diffusion_architecture}. Our code is accessible at the repository \href{https://github.com/zhonghl945/FDM_RTE_2D}{\textcolor{blue}{\url{https://github.com/zhonghl945/FDM_RTE_2D}}}. Unless otherwise stated, reconstruction accuracy is measured by the relative $L_2$ error over the prescribed target region:
\begin{equation}
\mathcal E_{\mathrm{rel}}
=
\frac{
\|\rho_{\mathrm{sol}}-\rho^*\|_{L_2([0,T_f]\times D^*)}
}{
\|\rho^*\|_{L_2([0,T_f]\times D^*)}
}.
\label{eq:numerical_relative_error}
\end{equation}

\subsection{Node-count Scan and Ablation Study}
\label{subsec4.1}

We first evaluate FDM+ in the homogeneous-medium configuration
described in \ref{app:rte_data}.

\begin{figure}[H]
    \centering
    \includegraphics[width=1.0\textwidth]{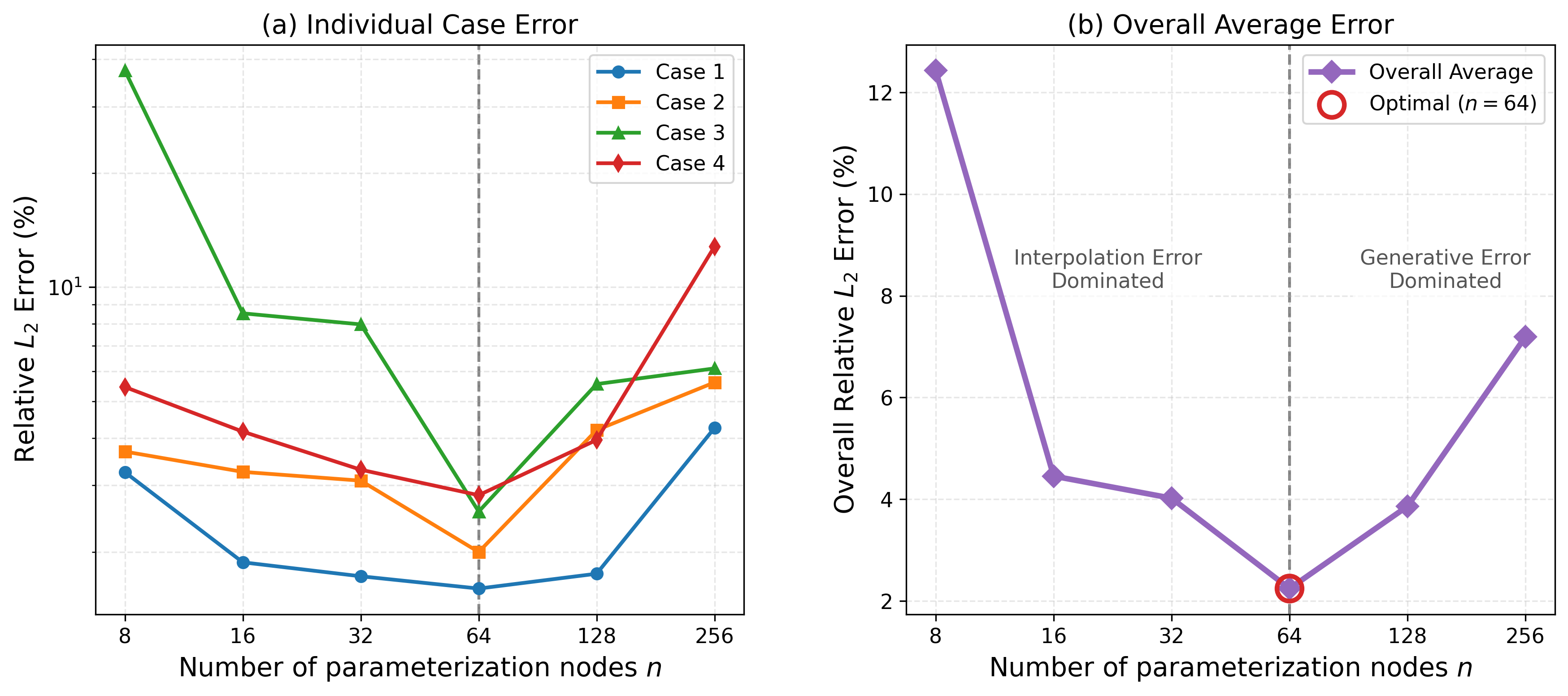}
    \caption{The effect of parameterization node count $n$ on the reconstruction error.}
    \label{fig:node_scan}
\end{figure}

On the one hand, we performed a node-count scan experiment by training FDM+ under different parameterization node counts: $n \in \{16, 32, 64, 128, 256\}$. For each configuration, we measured the reconstruction error on the test set. The experimental results are shown in \cref{fig:node_scan}. For low values of $n$, the interpolation error dominates, resulting in high reconstruction error. As $n$ increases, the error decreases significantly, reaching its minimum at $n=64$. However, as $n$ increases further, the generative estimation error starts to dominate because the diffusion model has to learn a much higher-dimensional distribution, which triggers the curse of dimensionality under limited training samples.

On the other hand, we conducted an ablation study to verify the individual contributions of the decoupled prediction and the control parameterization. It is crucial to note that in the previous node-count scan experiment, the low initial temperature ($T_0 = 10^{-3}$) introduces severe stiffness into the high-fidelity solver, forcing it to take thousands of adaptive time steps. Under such conditions, control parameterization is strictly necessary. To construct a feasible unparameterized baseline for this ablation study, we raised the initial temperature to $T_0 = 1.0$. This allows the traditional solver to run with fewer uniform time steps, making parameterization an option rather than a necessity. We select $n=128$ time nodes as our strictly unparameterized baseline.
\begin{itemize}
    \item \textbf{DC+P (Decoupled + Parameterized):} Direct value prediction on control, using $n=64$ nodes.
    \item \textbf{C+P (Coupled + Parameterized):} Standard noise prediction on control, using $n=64$ nodes.
    \item \textbf{DC+UP (Decoupled + Unparameterized):} Direct value prediction on control, using the full temporal resolution of $n=128$ time steps.
    \item \textbf{C+UP (Coupled + Unparameterized):} Standard noise prediction on control, using the full temporal resolution of $n=128$ time steps.
\end{itemize}

\begin{table}[H]
\centering
\caption{Reconstruction errors in ablation study. (DC: Decoupled, C: Coupled, P: Parameterized, UP: Unparameterized)}
\label{tab:ablation}
\footnotesize
\setlength{\tabcolsep}{3pt}
\begin{tabular}{lcccc}
\toprule
\textbf{Configuration} & \textbf{Case 1} & \textbf{Case 2} & \textbf{Case 3} & \textbf{Case 4} \\
\midrule
\textbf{DC+P (Ours)} & \textbf{0.95\% $\pm$ 0.60\%} & \textbf{0.64\% $\pm$ 0.22\%} & \textbf{0.35\% $\pm$ 0.09\%} & \textbf{1.54\% $\pm$ 0.82\%} \\
C+P & 145.60\% $\pm$ 0.45\% & 15.20\% $\pm$ 0.35\% & 30.80\% $\pm$ 0.08\% & 295.10\% $\pm$ 1.80\% \\
DC+UP & \underline{1.97\% $\pm$ 0.68\%} & \underline{0.82\% $\pm$ 0.30\%} & \underline{0.58\% $\pm$ 0.19\%} & \underline{3.43\% $\pm$ 0.67\%} \\
C+UP & 190.95\% $\pm$ 3.72\% & 20.15\% $\pm$ 0.41\% & 39.21\% $\pm$ 0.38\% & 409.23\% $\pm$ 5.35\% \\
\bottomrule
\addlinespace[3pt]
\multicolumn{5}{l}{\scriptsize \textit{Note:} Bold and underlined values indicate the lowest and second lowest errors, respectively.} \\
\end{tabular}
\end{table}

\cref{tab:ablation} highlights two key advantages of FDM+. First, decoupled prediction prevents the diffusion model from introducing additional spatial noise into the generated control, eliminating the massive errors observed in coupled models. Second, lowering the control nodes to a certain extent effectively alleviates the curse of dimensionality, thereby reducing reconstruction errors.

\subsection{Comparative Methods}
\label{subsec4.2}
To evaluate FDM+ under more challenging scenarios, we test the framework on two heterogeneous media situations and compare its performance against three baseline methods: DiffPhyCon \cite{wei2024diffphycon}, FNO \cite{li2021fourier}, and DeepONet \cite{lu2021learning}. The medium parameters, boundary
conditions, and test controls are reported in \ref{app:rte_data}.

\subsubsection{Central Strong-Absorption}
\label{subsubsec4.2.1}
\begin{figure}[H]
    \centering
    \includegraphics[width=0.7\textwidth]{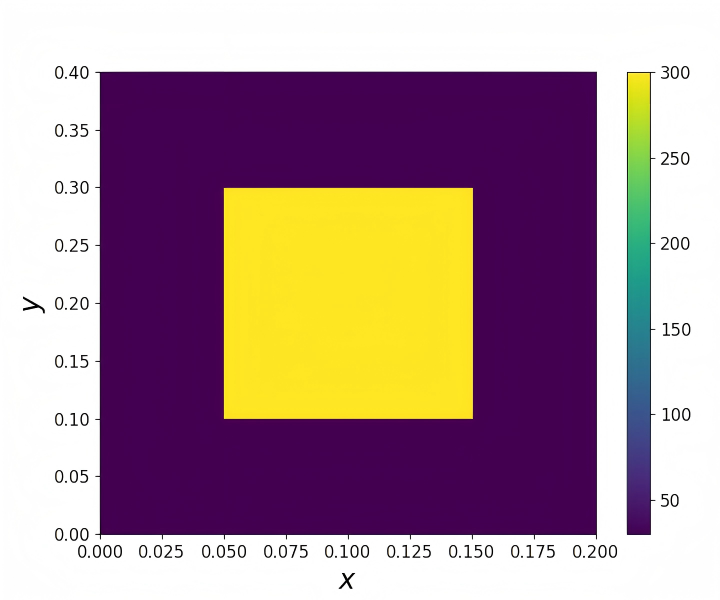} 
    \caption{The spatial distribution of the medium absorption coefficient $\sigma$ under central strong-absorption layered medium.}
    \label{fig:Multi_center}
\end{figure}

Consider a layered medium problem, illustrated in \cref{fig:Multi_center}. A rectangular layer with extremely high opacity is embedded in a transparent background medium. This abrupt, tenfold change in opacity significantly alters radiation transport paths, posing a substantial challenge to the control algorithm. The objective is to control the scalar flux within a vertical strip that traverses this high-opacity region.

\begin{figure}[H]
    \centering
    \includegraphics[width=0.7\textwidth]{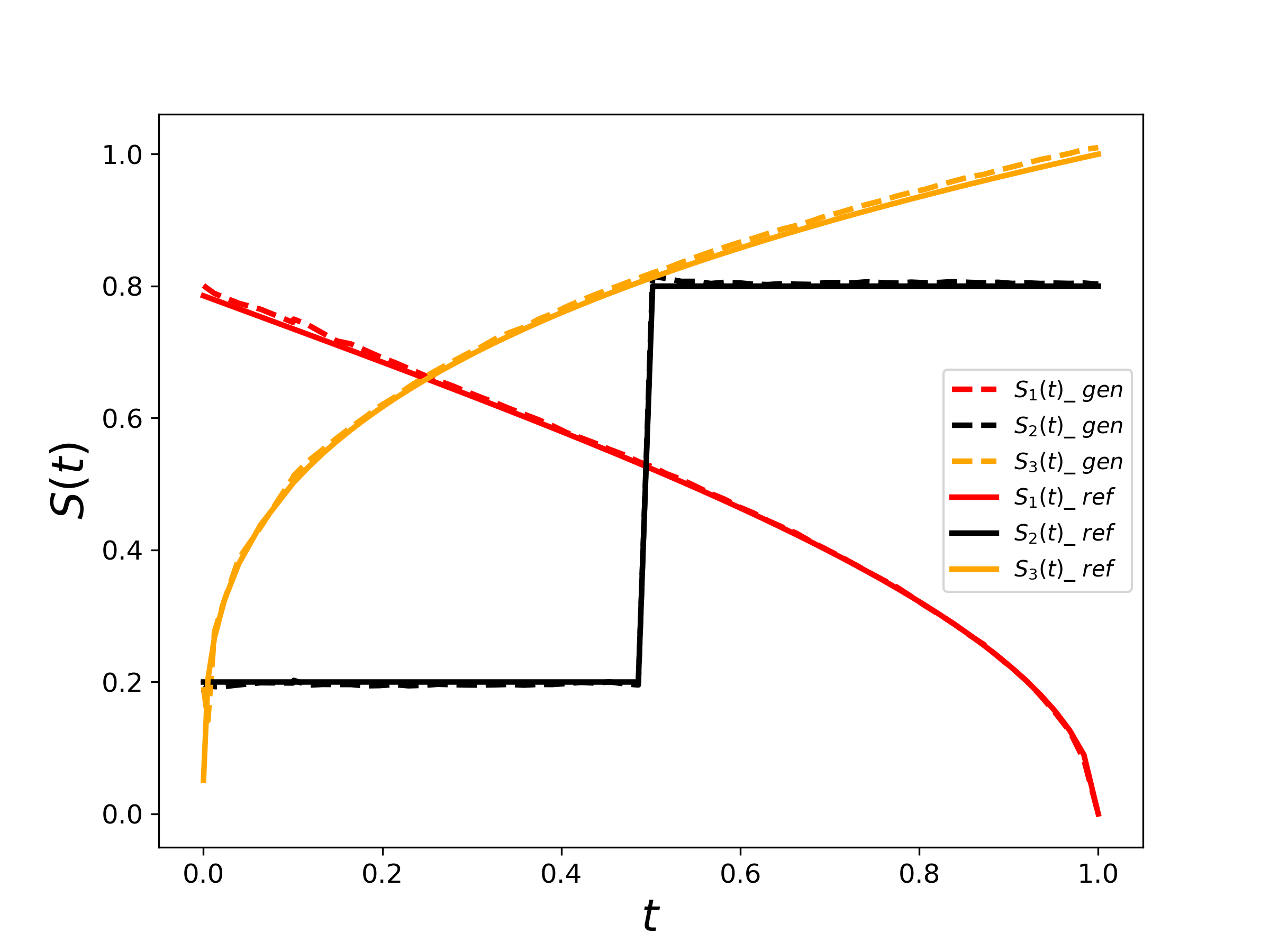} 
    \caption{Comparison of the generated $S(t)$ (dashed lines) and the ground truth (solid lines) in \cref{subsubsec4.2.1}.}
    \label{fig:Multi_S_center}
\end{figure}

\begin{figure}[H]
    \centering
    \includegraphics[width=1.0\textwidth]{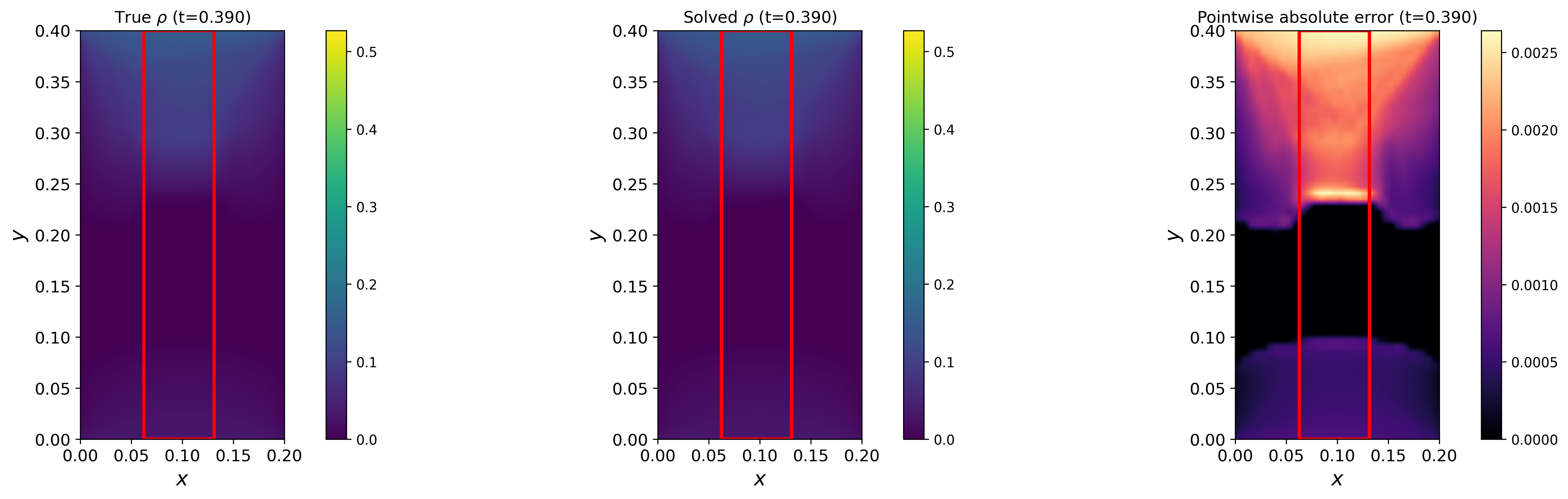}
    \includegraphics[width=1.0\textwidth]{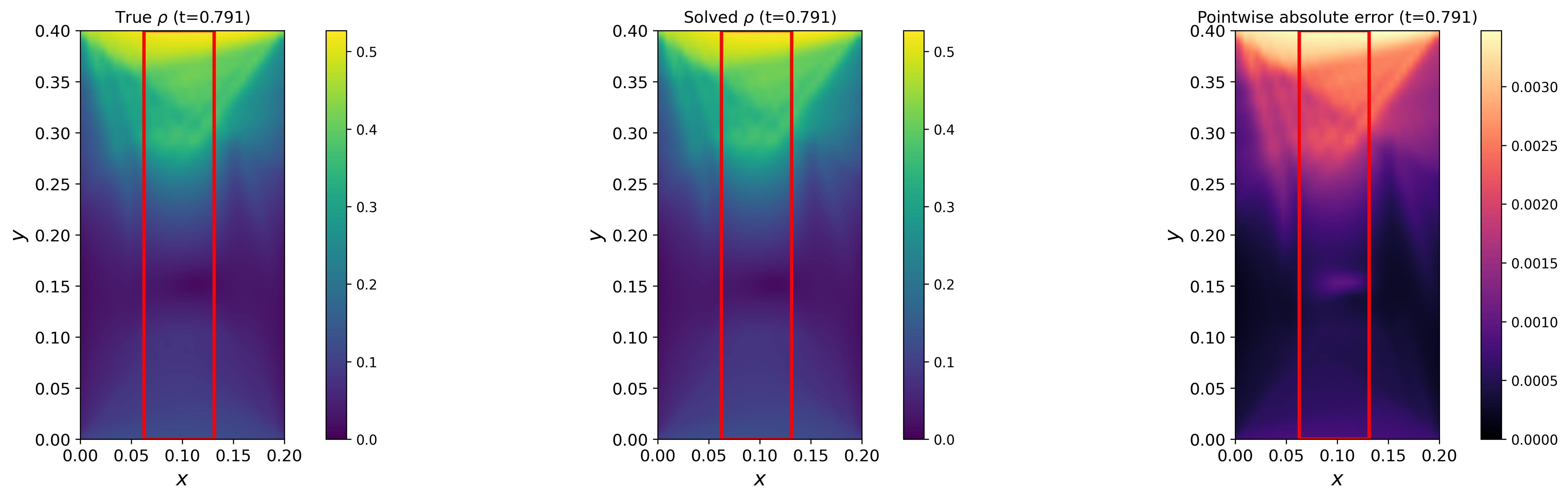}
    \includegraphics[width=1.0\textwidth]{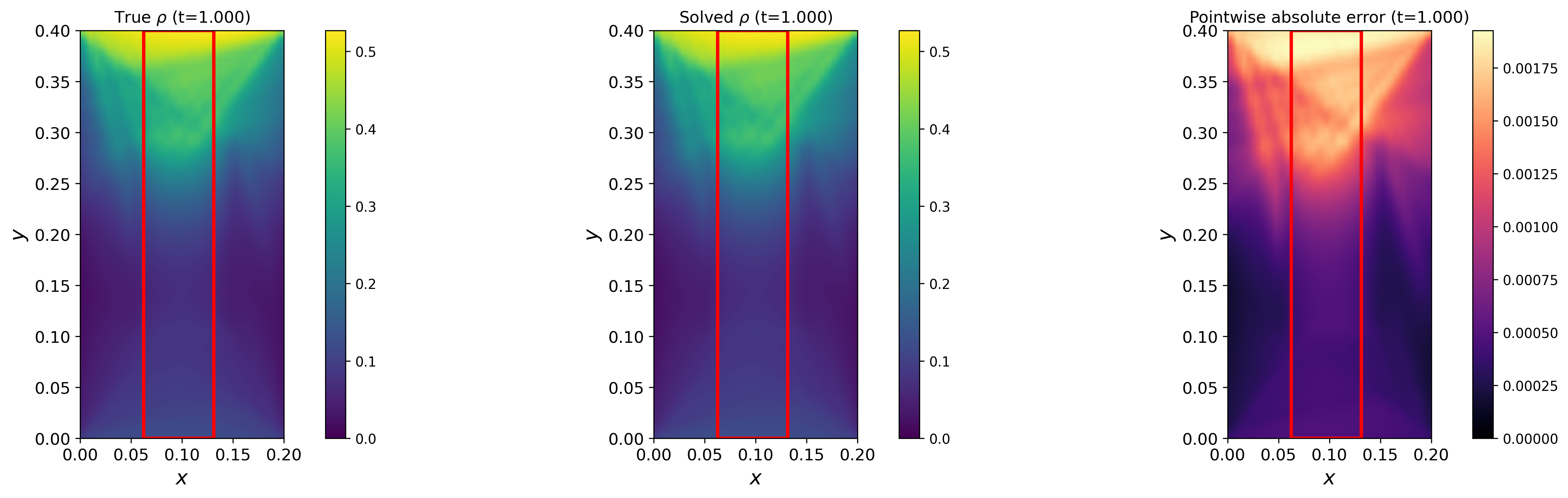}
    \caption{Reconstruction results for test case 2 under central strong-absorption layered medium at different times. \textbf{(Left)} Target scalar flux $\rho^*$; \textbf{(Middle)} Solved scalar flux $\rho_{\text{sol}}$; \textbf{(Right)} Pointwise absolute error $|\rho_{\text{sol}} - \rho^*|$. The red box indicates the constrained region $D^*$. The three rows correspond to t=0.390, 0.791, and 1.000, respectively.}
    \label{fig:Multi_rho_center}
\end{figure}

The results demonstrate that our method remains highly accurate even with sharp spatial variations in the medium. \cref{fig:Multi_S_center} illustrates the excellent agreement between the generated control functions $S_{gen}(t)$ and the ground truth. \cref{fig:Multi_rho_center} shows that when these controls are evaluated in the high-fidelity solver, the resulting flux fields accurately match the target states across almost all time steps. \cref{tab:compare_center} summarizes the comparative quantitative results.

\begin{table}[H]
\centering
\caption{Reconstruction errors under different methods in \cref{subsubsec4.2.1}.}
\label{tab:compare_center}
\small
\begin{tabular}{lccc}
\toprule
\textbf{Model} & \textbf{Case 1} & \textbf{Case 2} & \textbf{Case 3} \\
\midrule
\textbf{Ours (FDM+)} & \textbf{0.97\% $\pm$ 0.06\%} & \textbf{0.80\% $\pm$ 0.08\%} & \textbf{0.92\% $\pm$ 0.05\%} \\
DiffPhyCon & 2.45\% $\pm$ 0.15\% & 3.10\% $\pm$ 0.22\% & 2.78\% $\pm$ 0.18\% \\
FNO & 5.86\% & 7.94\% & 6.12\% \\
DeepONet & 4.88\% & 6.54\% & 7.35\% \\
\bottomrule
\end{tabular}%
\end{table}

\subsubsection{Tophat Maze}
\label{subsubsec4.2.2}
\begin{figure}[H]
    \centering
    \includegraphics[width=0.4\textwidth]{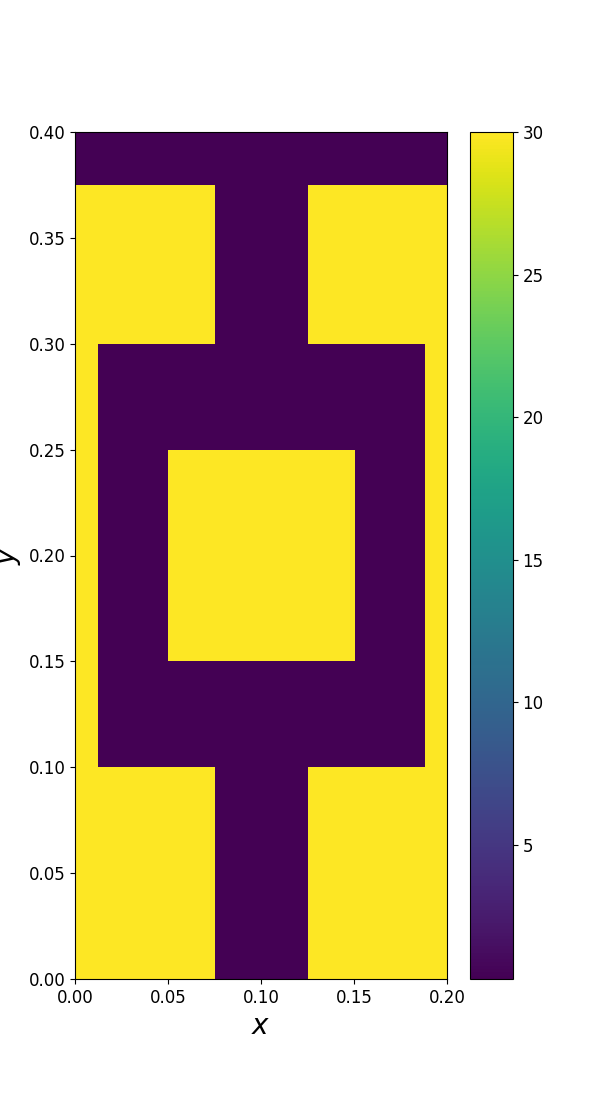} 
    \caption{The spatial distribution of the medium absorption coefficient $\sigma$ under Tophat Maze.}
    \label{fig:Multi_tophat}
\end{figure}

We further escalate the complexity of spatial heterogeneity by introducing a classic Tophat problem, illustrated in \cref{fig:Multi_tophat}. The medium features multiple optically thick obstacles forming a maze-like structure within an optically thin background. The control objective is to constrain the flux at the exit of the maze.

\begin{figure}[H]
    \centering
    \includegraphics[width=0.7\textwidth]{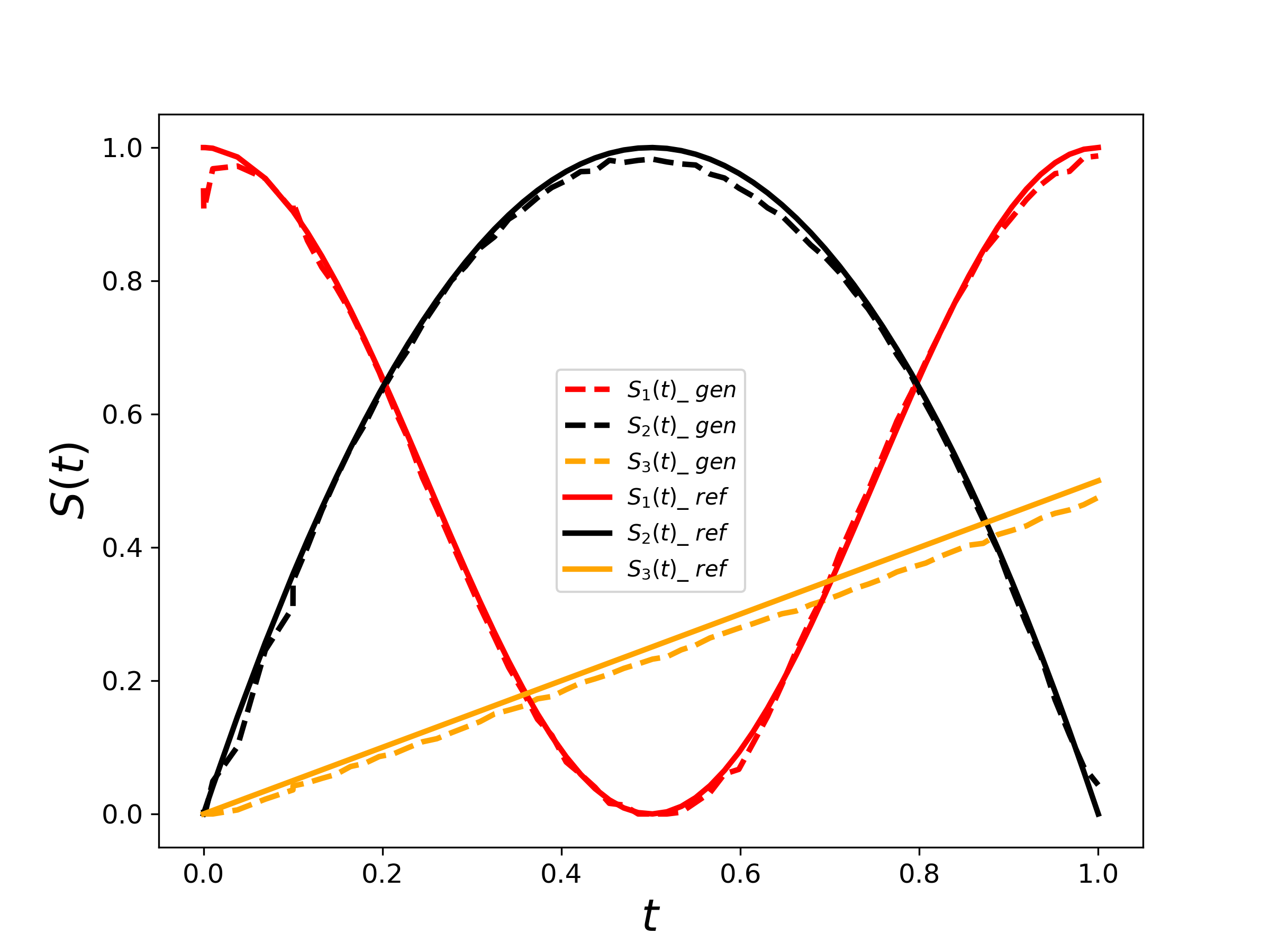} 
    \caption{Comparison of the generated $S(t)$ (dashed lines) and the ground truth (solid lines) in \cref{subsubsec4.2.2}.}
    \label{fig:Multi_S_tophat}
\end{figure}

\begin{figure}[H]
    \centering
    \includegraphics[width=1.0\textwidth]{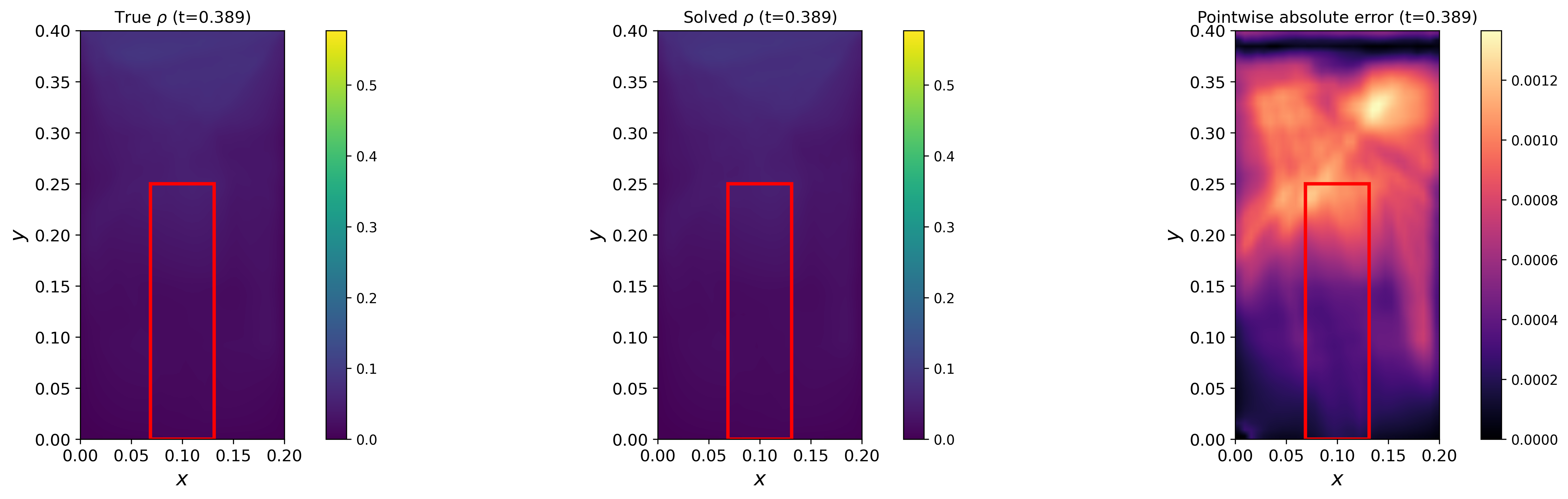}
    \includegraphics[width=1.0\textwidth]{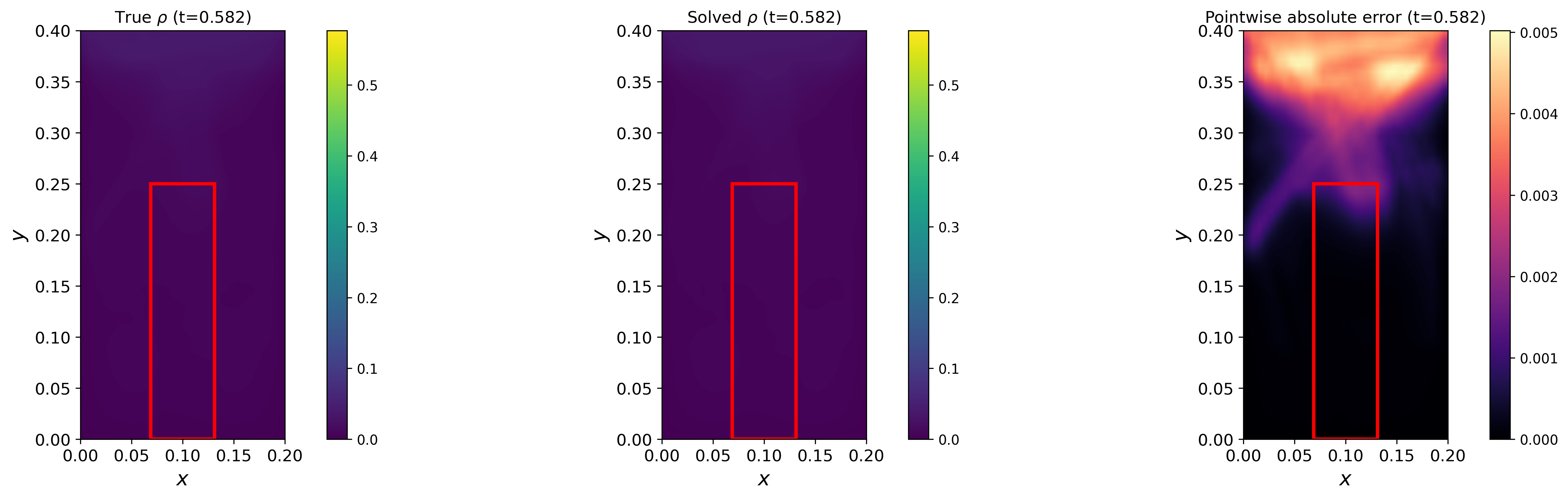}
    \includegraphics[width=1.0\textwidth]{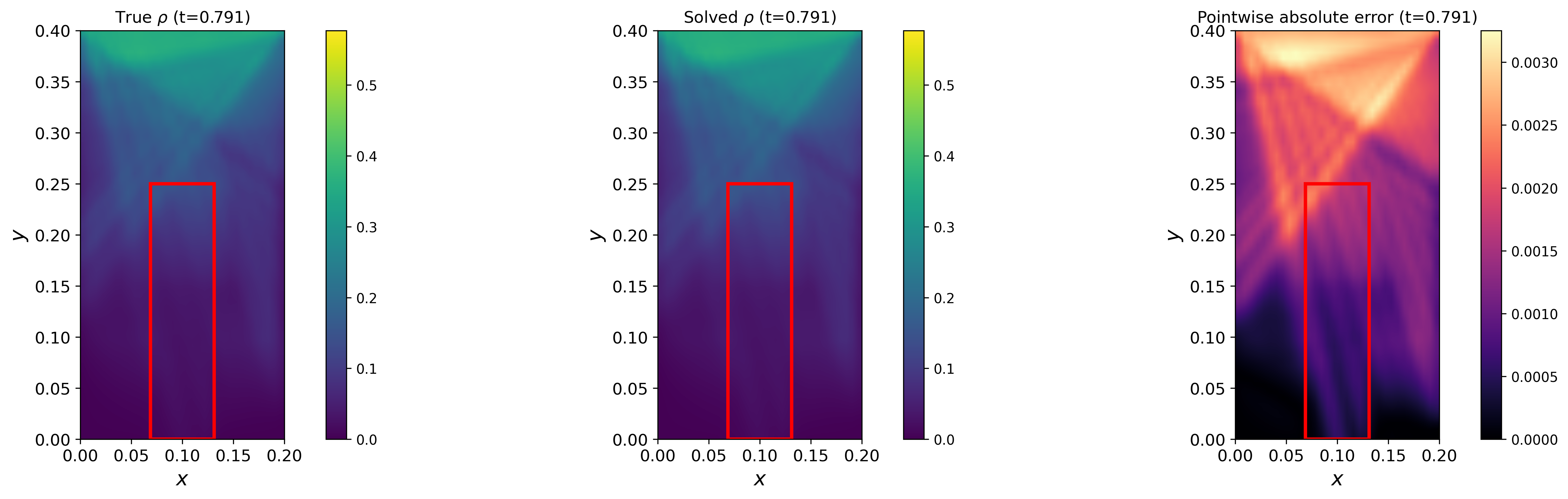}
    \caption{Reconstruction results for test case 1 under Tophat Maze at different times. \textbf{(Left)} Target scalar flux $\rho^*$; \textbf{(Middle)} Solved scalar flux $\rho_{\text{sol}}$; \textbf{(Right)} Pointwise absolute error $|\rho_{\text{sol}} - \rho^*|$. The red box indicates the constrained region $D^*$. The three rows correspond to t=0.390, 0.582, and 0.791, respectively.}
    \label{fig:Multi_rho_tophat}
\end{figure}

Even with this significant increase in geometric complexity, the model demonstrates powerful and robust performance. \cref{fig:Multi_S_tophat} reveals a close match between the generated boundary controls and the ground truth. Furthermore, as shown in \cref{fig:Multi_rho_tophat}, these controls successfully drive the radiation through the complex maze to produce the target state at the designated exit. Finally, \cref{tab:compare_tophat} summarizes the comparative quantitative results.

\begin{table}[H]
\centering
\caption{Reconstruction errors under different methods in \cref{subsubsec4.2.2}.}
\label{tab:compare_tophat}
\small
\begin{tabular}{lccc}
\toprule
\textbf{Model} & \textbf{Case 1} & \textbf{Case 2} & \textbf{Case 3} \\
\midrule
\textbf{Ours (FDM+)} & \textbf{1.72\% $\pm$ 0.12\%} & \textbf{2.19\% $\pm$ 0.15\%} & \textbf{2.90\% $\pm$ 0.18\%} \\
DiffPhyCon & 4.12\% $\pm$ 0.25\% & 4.85\% $\pm$ 0.32\% & 5.24\% $\pm$ 0.28\% \\
FNO & 8.65\% & 12.30\% & 9.78\% \\
DeepONet & 10.42\% & 9.85\% & 11.60\% \\
\bottomrule
\end{tabular}%
\end{table}

\subsection{Multiple Solution}
\label{subsec4.3}

In highly scattering or optically thick regimes, the RTEs asymptotically converges to the diffusion limit. To demonstrate the capability of FDM+ in exploring the design space under this regime, we investigate a 2D nonlinear diffusion problem as a representative physical model. The state evolution is governed by:
\begin{equation}
    \frac{\partial T}{\partial t} - \nabla \cdot \left(\kappa(T, x, y) \nabla T \right) = F(x, y),
\end{equation}
where the temperature-dependent thermal conductivity is:
\begin{equation}
    \kappa(T, x, y) = \frac{T^3}{k_0(x,y)}, \quad \text{where } k_0(x,y) = 
    \begin{cases}
        50 & \text{if } (x,y) \in [0.25, 0.75]^2, \\
        300 & \text{otherwise}.
    \end{cases}
\end{equation}
The system evolves in the domain $D=[0,1]^2$ over $t \in [0, 0.5]$, with an initial temperature of $T(0,x,y)=0.1$, which is also held constant on all boundaries. Here, the state variable is the temperature field $T(t,x,y)$, and the control variable is the time-independent spatial heat source $F(x,y)$. The control objective is to reconstruct a source $F^*(x,y)$ such that the final temperature field $T(t=0.5, x, y)$ matches a target $T^*$ within a central constraint region $D^* = [0.4, 0.6]^2$. Further details on the training dataset and training hyperparameters are provided in \ref{app:rde_data}.

\begin{figure}[H]
    \centering
    \includegraphics[width=1.0\textwidth]{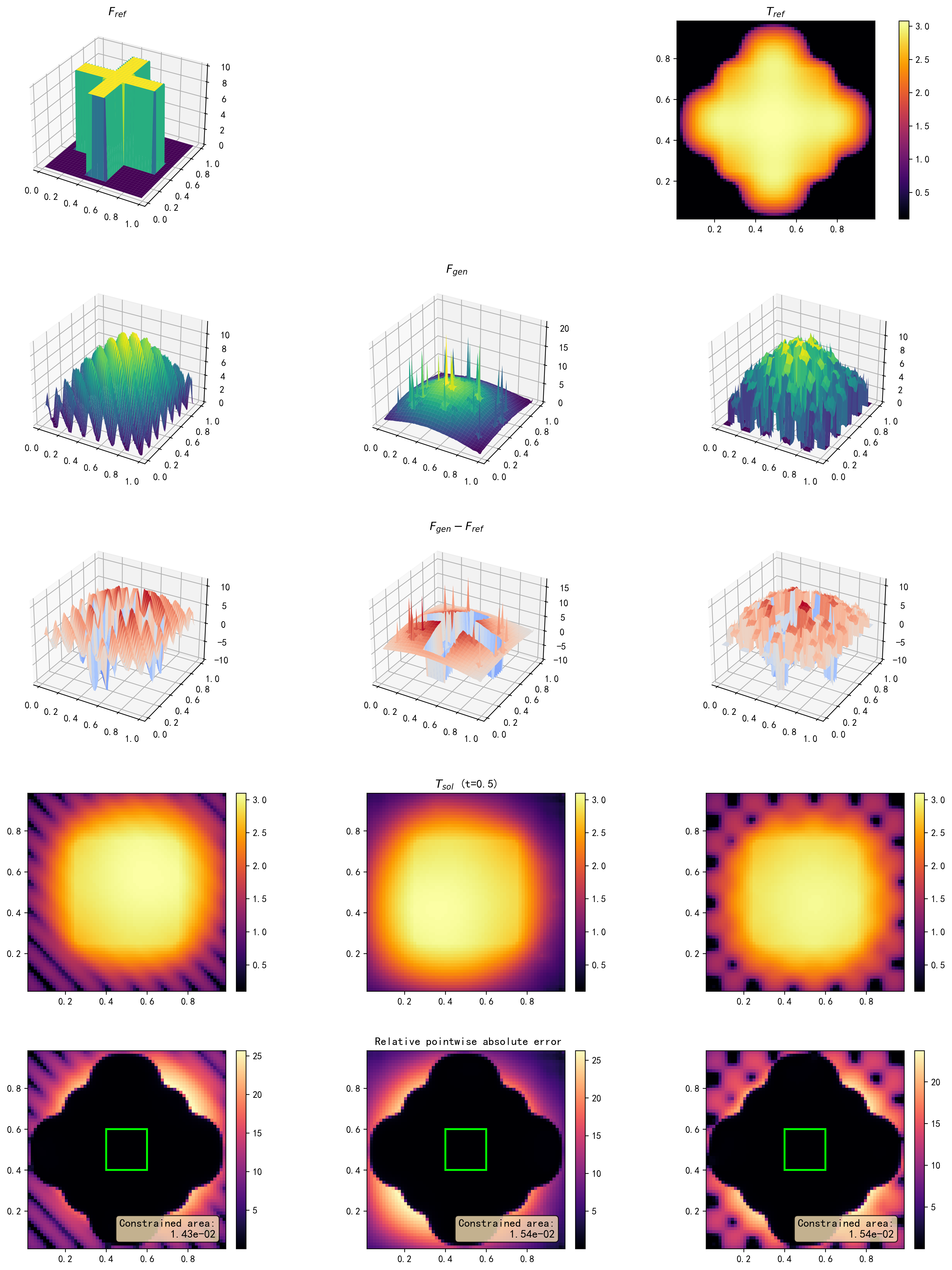}
    \caption{Visualization of the nulti-solution. \textbf{(Row 1)} The reference source $F_{ref}$ and its resulting target temperature $T_{ref}$. \textbf{(Rows 2-4)} The generated source term $F_{gen}$, source difference $F_{gen} - F_{ref}$, and the final solved temperature field $T_{sol}$ for three representative samples. \textbf{(Row 5)} The relative pointwise absolute error maps. The green box indicates the constraint region $D^*$, with the average error for this region shown below.}
    \label{fig:Multiple_solution}
\end{figure}

We verify this multi-solution capability by generating an ensemble of 50 candidate source terms $\{F_{gen,i}\}$. \cref{fig:Multiple_solution} visually confirms that FDM+ reconstructs multiple, morphologically diverse solutions that are clearly distinct from each other and the reference source $F_{ref}$. Despite their different forms, each generated source produces a final temperature field that is nearly identical to the target within the constrained region (green box), with errors on the order of 1\%.

\begin{figure}[H]
    \centering
    \includegraphics[width=1.0\textwidth]{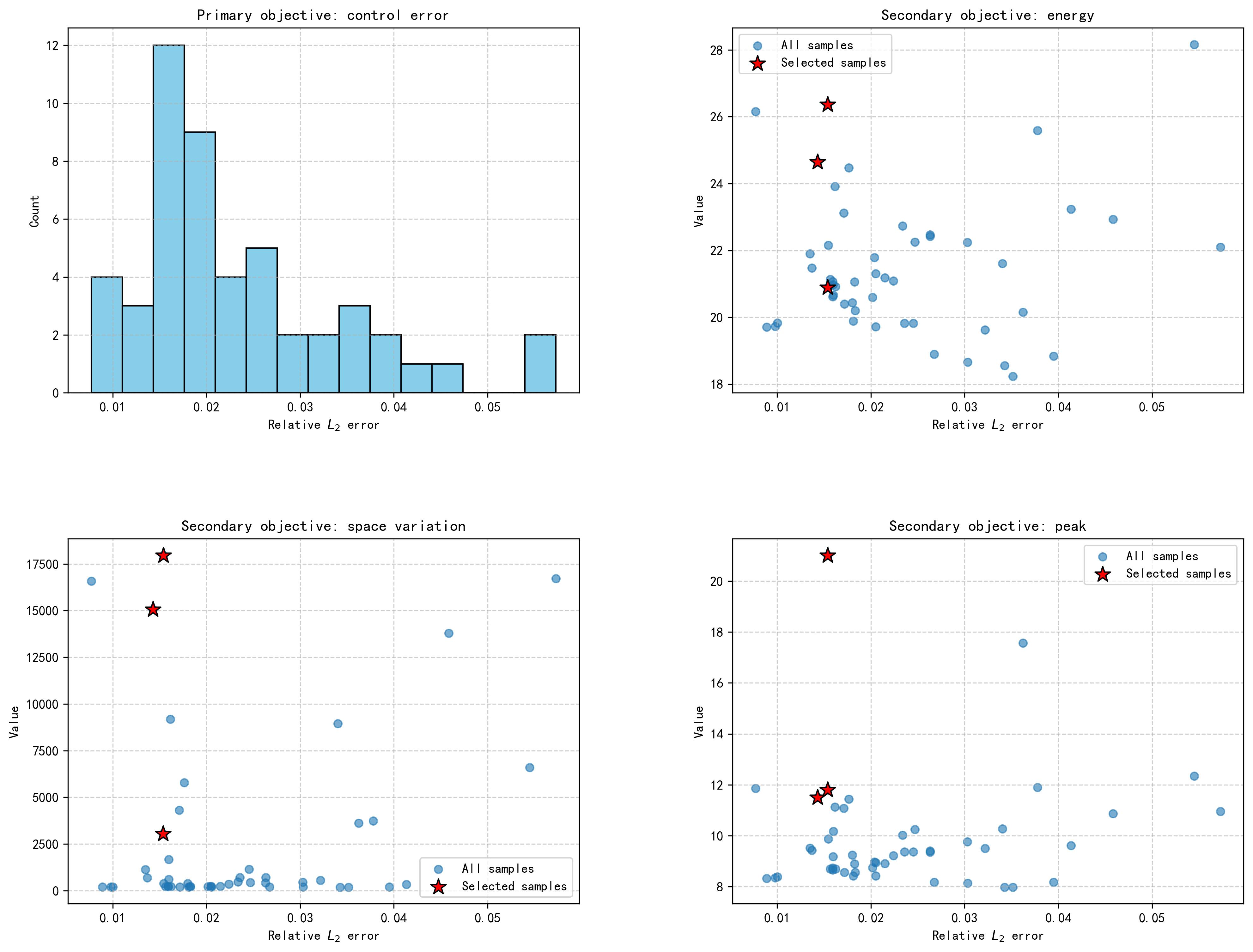}
    \caption{Multi-objective performance analysis of the 50-sample ensemble. \textbf{(Top-left)} Histogram of the primary control error. \textbf{(Others)} Scatter plots showing the trade-offs between the primary error and three secondary objectives (energy, space variation, and peak value). Star markers indicate the three representative samples from \cref{fig:Multiple_solution}.}
    \label{fig:Multiple_target}
\end{figure}

Furthermore, this multi-solution capability allows designers to perform a posteriori design selection. For this analysis, we define three secondary objectives for a given source term $F(x,y)$ over the domain $D$ as follows:
\begin{itemize}
    \item \textbf{Energy:} The total power of the source, defined as the $L_2$ norm squared:
    \begin{equation}
        J_{\text{energy}}(F) = \int_D F(x,y)^2 \,dx\,dy.
    \end{equation}
    \item \textbf{Space variation:} A measure of the source's spatial roughness, defined by the squared norm of its gradient:
    \begin{equation}
        J_{\text{space\_variation}}(F) = \int_D ||\nabla F(x,y)||^2 \,dx\,dy.
    \end{equation}
    \item \textbf{Peak:} The maximum absolute intensity of the source, defined by the $L_\infty$ norm:
    \begin{equation}
        J_{\text{peak}}(F) = \max_{(x,y) \in D} |F(x,y)|.
    \end{equation}
\end{itemize}
\cref{fig:Multiple_target} shows the multi-objective performance analysis of the 50-sample ensemble. The error histogram confirms the model's high fidelity, with most of the 50 samples clustering at a low error. More importantly, the scatter plots reveal a critical trade-off issue, analogous to the Pareto fronts sought by traditional multi-objective optimization techniques \cite{coello2007evolutionary}. Even for solutions with similarly low primary control errors, their performance on secondary metrics like energy and smoothness varies widely. This provides designers with significant latitude for a posteriori decision-making, allowing them to select a solution that best fits their specific engineering needs without sacrificing reconstruction accuracy.

\section{Conclusion}
\label{sec5}

In this work, we developed FDM+, a deep learning framework for the constrained control of 2D radiative transport equation (RTE). Through a decoupled prediction architecture, a low-dimensional control parameterization, and an a posteriori multi-objective selection strategy, FDM+ addresses the challenges of dimensional heterogeneity, scale mismatch, and solution non-uniqueness. On the theoretical side, we established a rigorous total error bound for the parameterized control representation. The bound separates the interpolation error from the statistical distribution-estimation error and clarifies how the node count influences the reconstruction performance. Numerically, FDM+ demonstrates significant and consistent quantitative advantages. Our decoupled and parameterized architecture effectively eliminates the massive errors (>100\%) inherent to baseline designs, reducing to 0.35\%–1.54\%. Across various complex heterogeneous media, FDM+ consistently maintains sub-3\% reconstruction errors, delivering a 2$\times$ to 8$\times$ error reduction compared to other methods such as DiffPhyCon, FNO, and DeepONet. Furthermore, the framework can generate diverse candidate controls with primary errors around 1\%, enabling flexible a posteriori selection to balance secondary engineering objectives without sacrificing accuracy.

Several limitations remain for future work. First, the current framework assumes that the background medium is known and stationary. Extending FDM+ to dynamic or uncertain media would improve its robustness in more realistic physical environments. Second, the present study focuses on 2D settings with isotropic physics. Scaling the framework to 3D problems with anisotropic scattering will require more advanced architectures and more efficient parallel computation. These directions will further broaden the applicability of FDM+ to high-dimensional engineering control problems.

\section*{Code availability}
The code is accessible at \href{https://github.com/zhonghl945/FDM_RTE_2D}{https://github.com/zhonghl945/FDM\textunderscore RTE\textunderscore 2D}.

\section*{Acknowledgments}
The authors sincerely appreciate the support and discussions provided by all the members from YONG HENG’s team “AI++”, as well as Professor Zhiping Mao at Eastern Institute of Technology, Ningbo. The work is partially supported by the Science Challenge Project (No.TZ2025008), National Natural Science Foundation of China (NSFC) under Grant Nos. (12331010, 12371390) and the National Safety Academic Fund (NSAF) under Grant No. U2230208.

\section*{Declaration of competing interest}
The authors declare that they have no known competing financial interests or personal relationships that could have appeared to influence the work reported in this paper.

\section*{Declaration of generative AI and AI-assisted technologies in the manuscript preparation process}
During the preparation of this work, the author(s) used Google Gemini in order to improve the language and readability of the manuscript. After using this tool/service, the author(s) reviewed and edited the content as needed and take(s) full responsibility for the content of the published article.

\bibliographystyle{elsarticle-num}

\appendix

\section{Data Generation}
\label{app:data_generation}

\subsection{Radiative Transport Equation}
\label{app:rte_data}

\subsubsection{Training Data}
The training data for the 2D RTE were generated using a high-fidelity solver based on the $S_N$ diamond-difference scheme \cite{lathrop1969spatial}. The spatial domain $D=[0,0.2]\times[0,0.4]$ was discretized on a $33\times65$ grid. An $S_8$ Gauss--Legendre
quadrature was used for the angular discretization, resulting in
64 discrete directions. In all configurations, the control
$S(t)$ was imposed at the top boundary $y=0.4$, while the left and
right boundaries had zero inflow.

\paragraph{Homogeneous medium}
The absorption coefficient was spatially uniform with
$\sigma=100$. The temporal interval was $[0,0.5]$, the bottom
boundary condition was $0.2S(t)$, and the target region was
$D^*=[0,0.2]\times[0.1,0.3]$.

For the node scan experiment, the initial temperature was
$T_0=10^{-3}$ and adaptive time integration was used to handle the
resulting stiffness. Separate models were trained with
$n\in\{16,32,64,128,256\}$.

For the ablation study, the initial temperature was increased to
$T_0=1$, and the temporal interval was discretized using 128 uniform steps. The parameterized models used $n=64$.

\paragraph{Central strong absorption medium}
The temporal interval was $[0,1]$, and adaptive time integration
produced approximately 3,046 solver steps. The initial temperature
was $T_0=10^{-6}$. A rectangular region with $\sigma=300$ was embedded in a background with $\sigma=30$. The bottom boundary condition was $0.2S(t)$, and the target region was
$D^*=[0.0625,0.13125]\times[0,0.4]$.

\paragraph{Tophat maze medium}
The temporal interval and initial condition were the same as those
of the central strong-absorption configuration. Optically thick
obstacles with $\sigma=30$ formed a maze within a background with
$\sigma=0.3$. The bottom boundary had zero inflow, and $D^*$ was
located at the exit of the maze.

Each RTE configuration contained 1,000 training trajectories. For
each trajectory, a boundary control was sampled and evaluated by the high-fidelity solver to produce a paired control--state sample. Let $\tau=t / T_f\in[0,1]$ denote normalized time, where $T_f$ is the final time of the corresponding configuration. The training controls were sampled from the following eight families:
\begin{equation}
\label{eq:rte_training_controls}
\left\{
\begin{aligned}
S_{\mathrm{sin}}(\tau)
    &=|\sin(4\pi k_1\tau)|,
    &&k_1\sim U[0,1],\\
S_{\mathrm{exp}}(\tau)
    &=\exp\!\bigl(a(\tau-c)\bigr),
    &&a\sim U[-1,1],\quad c=\mathbb I(a>0),\\
S_{\mathrm{lin}}(\tau)
    &=|a(\tau-c)|,
    &&a\sim U[-1,1],\quad c=\mathbb I(a<0),\\
S_{\mathrm{quad}}(\tau)
    &=(\tau-k_1)^2,
    &&k_1\sim U[0,1],\\
S_{\mathrm{cubic}}(\tau)
    &=\frac{1}{3}|\tau-k_1|^3+\frac{2}{3}k_2,
    &&k_1,k_2\sim U[0,1],\\
S_{\mathrm{atan}}(\tau)
    &=\frac{2}{3}\arctan(k_1\tau),
    &&k_1\sim U[0,20],\\
S_{\mathrm{gauss}}(\tau)
    &=\exp\!\left[-\frac{(\tau-\mu)^2}{2\sigma_r^2}\right],
    &&\mu\sim U[0,1],\quad \sigma_r\sim U(0,0.2],\\
S_{\mathrm{rat}}(\tau)
    &=\frac{|k_1|}{k_2\tau+2},
    &&k_1,k_2\sim U[-1,1].
\end{aligned}
\right.
\end{equation}
Each sampled function was multiplied by an independent amplitude
factor drawn from $U[0.5,2.0]$.

\subsubsection{Test Data}
The RTE test controls included both in-distribution (ID) of training
families and out-of-distribution (OOD) functional forms. For each test control, the target scalar-flux trajectory was generated using the same high-fidelity solver as in the training-data construction.

\begin{table}[H]
\centering
\caption{Control functions for the homogeneous medium test set.}
\label{tab:homogeneous_test_controls}
\small
\begin{tabular}{lll}
\toprule
\textbf{Case} & \textbf{Type} & \textbf{Control function} \\
\midrule
1 & OOD & $S(\tau)=1.8(2^\tau-1)$ \\
2 & OOD & $S(\tau)=-\log(0.5\tau+0.3)$ \\
3 & ID & $S(\tau)=(\tau-k_1)^2$ \\
4 & ID & $S(\tau)=\frac13|\tau-k_1|^3+\frac23k_2$ \\
\bottomrule
\end{tabular}
\end{table}

\begin{table}[H]
\centering
\caption{Control functions for the heterogeneous medium test sets.}
\label{tab:heterogeneous_test_controls}
\small
\begin{tabular}{lll}
\toprule
\textbf{Configuration} & \textbf{Case} & \textbf{Control function} \\
\midrule
Central strong absorption
  & 1 & $S(\tau)=0.5\arccos(\tau)$ \\
  & 2 &
  $S(\tau)=
  \begin{cases}
  0.2, & \tau<0.5,\\
  0.8, & \tau\geq0.5
  \end{cases}$ \\
  & 3 & $S(\tau)=\tau^{0.3}$ \\
\midrule
Tophat maze
  & 1 & $S(\tau)=0.5(1+\cos(2\pi\tau))$ \\
  & 2 & $S(\tau)=4\tau(1-\tau)$ \\
  & 3 & $S(\tau)=0.5\tau$ \\
\bottomrule
\end{tabular}
\end{table}

\subsection{Nonlinear Diffusion Equation}
\label{app:rde_data}

\subsubsection{Training Data}
The nonlinear diffusion equation data were generated
using an explicit finite-difference solver on
$D=[0,1]\times[0,1]$. The domain was discretized on a
$66\times66$ grid, and 40,000 internal time steps were used to evolve
the system to $T_f=0.5$. For each sample, 16 temporal snapshots of
the temperature field $T(t,x,y)$ and the corresponding time-independent
source $F(x,y)$ were retained. The training set contained 50,000
samples.

Each source was formed as
\begin{equation}
F(x,y)=F_{\mathrm{base}}(x,y)+F_{\mathrm{pert}}(x,y),
\label{eq:rde_source_decomposition}
\end{equation}
where
\begin{equation}
F_{\mathrm{base}}(x,y)
=
A\exp\!\left(
-\frac{(x-c_x)^2+(y-c_y)^2}{0.2}
\right),
\label{eq:rde_base_source}
\end{equation}
with $A\sim U[5,15]$ and
$(c_x,c_y)\sim U([0.4,0.6]^2)$. With probability $0.7$,
$F_{\mathrm{pert}}$ was selected uniformly from the following six
families; otherwise, $F_{\mathrm{pert}}=0$:
\begin{enumerate}
    \item \textbf{Sinusoidal:}
    \[
    F_{\mathrm{pert}}(x,y)
    =
    A\sin(k_x\pi x)\sin(k_y\pi y),
    \]
    where $A\sim U[0,5]$ and $k_x,k_y\sim U[7,16]$.

    \item \textbf{Checkerboard:}
    \[
    F_{\mathrm{pert}}(x,y)
    =
    A(-1)^{\lfloor kx\rfloor+\lfloor ky\rfloor},
    \]
    where $A\sim U[1,5]$ and
    $k\in\{10,11,12,13,14\}$.

    \item \textbf{Perlin noise:}
    \[
    F_{\mathrm{pert}}(x,y)=A\,P(x,y),
    \]
    where $A\sim U[2,8]$ and $P$ is multi-octave Perlin noise
    with 4--9 octaves.

    \item \textbf{Diagonal waves:}
    \[
    F_{\mathrm{pert}}(x,y)
    =
    A\sin\!\bigl(k\pi(x+dy)\bigr),
    \]
    where $A\sim U[2,5]$, $k\sim U[7,15]$, and
    $d\in\{-1,1\}$.

    \item \textbf{Concentric waves:}
    \[
    F_{\mathrm{pert}}(x,y)=A\sin(kR),
    \qquad
    R=\sqrt{(x-c_x)^2+(y-c_y)^2},
    \]
    where $A\sim U[1,3]$, $k\sim U[20,30]$, and the center
    $(c_x,c_y)$ was sampled uniformly in $D$.

    \item \textbf{Sparse impulses:}
    a grid-based sparse impulse field with amplitudes sampled from
    $[-10,-5]\cup[5,10]$ and sparsity sampled between
    $0.5\%$ and $5\%$.
\end{enumerate}

\subsubsection{Test Data}
A single fixed cross-shaped source centered in the domain was used to construct the test target:
\begin{equation}
F_{\mathrm{ref}}(x,y)
=
A_{\mathrm{ref}}\,
\mathbb I\!\left(
|x-\tfrac12|\leq w
\ \text{or}\
|y-\tfrac12|\leq w
\right),
\label{eq:rde_reference_source}
\end{equation}
where $A_{\mathrm{ref}}$ and $w$ are the fixed amplitude and arm
half-width used in the numerical implementation. Solving $T(t,x,y)$
with $F_{\mathrm{ref}}$ produced the target temperature trajectory
$T^*$. The design-space experiment conditioned on this target and
generated 50 candidate sources using independent noise seeds.

\section{Network Architecture and Training}
\label{app:diffusion_architecture}

All experiments used a three-dimensional (3D) U-Net denoiser consisting of a downsampling encoder, a bottleneck, and an upsampling decoder with skip connections. The encoder contained three resolution levels with channel multipliers $(1,2,4)$ applied to an initial width of 64. Each level used ResNet blocks, and the diffusion time was represented by a 256-dimensional embedding.

The decoupled prediction mechanism used two output heads. For the RTE experiments, a 3D convolutional head predicted the
noise associated with the scalar-flux trajectory, while a second head combined adaptive average pooling and a 1D convolution to predict the $n$ control-node values directly. Thus, the output dimension of the control head was adjusted to the node count used in
each node-scan or ablation configuration. For the multi-solution experiment, the field head predicted the noise associated with the temperature trajectory, while the direct-prediction head produced the 2D source $F(x,y)$. The remaining settings are listed in \cref{tab:common_hyperparameters}.

\begin{table}[H]
\centering
\caption{Network and diffusion hyperparameters shared by all experiments.}
\label{tab:common_hyperparameters}
\small
\begin{tabular}{ll}
\toprule
\textbf{Hyperparameter} & \textbf{Value} \\
\midrule
\multicolumn{2}{c}{\textit{U-Net architecture}} \\
Initial channel dimension & 64 \\
Channel multipliers & $(1,2,4)$ \\
ResNet block groups & 8 \\
Time-embedding dimension & 256 \\
Padding mode & Reflect \\
\midrule
\multicolumn{2}{c}{\textit{Diffusion process}} \\
Diffusion steps & 1,000 \\
Beta schedule & Cosine \\
Direct-prediction loss weight $\lambda$ & 1.0 \\
\midrule
\multicolumn{2}{c}{\textit{Optimization}} \\
Batch size & 8 \\
Optimizer & Adam \\
Learning rate & $10^{-4}$ \\
Learning-rate scheduler & Cosine annealing with warmup \\
Warmup steps & 5,000 \\
EMA decay & 0.995 \\
Numerical precision & FP16 \\
Optimization steps & 100,000 \\
\bottomrule
\end{tabular}
\end{table}

\section{Proof}
\label{app:lipschitz_interpolation}

\begin{proof}[Proof of \cref{lem:lipschitz_interpolation}]
Let
\[
h=\frac{T_f}{n-1}.
\]
For $t\in[t_i,t_{i+1}]$, set
\[
\theta=\frac{t-t_i}{h}\in[0,1].
\]
Since
\[
S_{\mathrm{interp}}(t;\mathbf{s})
=
(1-\theta)S(t_i)+\theta S(t_{i+1}),
\]
the Lipschitz continuity of $S$ gives
\begin{align}
|S(t)-S_{\mathrm{interp}}(t;\mathbf{s})|
&\leq
(1-\theta)|S(t)-S(t_i)|
+
\theta|S(t)-S(t_{i+1})|
\nonumber\\
&\leq
2Lh\,\theta(1-\theta)
\leq
\frac{Lh}{2}.
\label{eq:appendix_pointwise_interpolation_bound}
\end{align}
Therefore,
\begin{equation}
\left\|
S-S_{\mathrm{interp}}(\cdot;\mathbf{s})
\right\|_{L_2}
\leq
\sqrt{T_f}\,\frac{Lh}{2}
=
\frac{LT^{3/2}_f}{2(n-1)}
\label{eq:appendix_l2_interpolation_bound}
\end{equation}
\end{proof}

\end{document}